%% file: emnlp2023.tex
\pdfoutput=1

\documentclass[11pt]{article}

\usepackage{EMNLP2023}

\usepackage{times}
\usepackage{latexsym}

\usepackage[T1]{fontenc}

\usepackage[utf8]{inputenc}

\usepackage{microtype}

\usepackage{inconsolata}

\usepackage{graphicx}
\usepackage{xcolor}
\usepackage{soul}
\newcommand{\hlc}[2][red]{{%
    \colorlet{foo}{#1}%
    \sethlcolor{foo}\hl{#2}}%
}
\usepackage{setspace}

\title{KL-Divergence Guided Temperature Sampling}

\author{Chung-Ching Chang \and David Reitter \and Renat Aksitov \and Yun-Hsuan Sung \\
        Google Research}

\begin{document}
\maketitle
\begin{abstract}
Temperature sampling is a conventional approach to diversify large language model predictions. As temperature increases, the prediction becomes diverse but also vulnerable to hallucinations -- generating tokens that are sensible but not factual. One common approach to mitigate hallucinations is to provide source/grounding documents and the model is trained to produce predictions that bind to and are attributable to the provided source. It appears that there is a trade-off between diversity and attribution. To mitigate any such trade-off, we propose to relax the constraint of having a fixed temperature over decoding steps, and a mechanism to guide the dynamic temperature according to its relevance to the source through KL-divergence. Our experiments justifies the trade-off, and shows that our sampling algorithm outperforms the conventional top-k and top-p algorithms in conversational question-answering and summarization tasks.
\end{abstract}

\section{Introduction}

The generative large language models (LLMs) have enabled many applications in conversations \citep{thoppilan2022lamda}, code generations \citep{chen2021evaluating}, and art creations \citep{yu2022scaling}. However, LLMs are known to hallucinate. While hallucinations may be fine for  creative tasks, \emph{e.g.} storytelling and art creations, it is not tolerable for other tasks, \emph{e.g.} fact-seeking question-answering (QA) and text summarization.

There are a few reasons why LLMs hallucinate. One reason is that they are trained on massive datasets of text, which can contain errors and biases. Furthermore, if the model is asked to predict when the input text pattern is not supported by the training data distributions, it may therefore hallucinate information to complete the prediction plausibly.

To overcome the limitation in the training data, an alternative is to incorporate LLMs with an external trusted knowledge source, such as search engines \citep{thoppilan2022lamda} and retrieval systems \citep{borgeaud2022improving}. These alternative approaches with an external trusted knowledge source transforms a factuality problem into query generation and source grounding problems. We call the setting of generating responses with source/grounding documents provided as \emph{contextual predictions}. 

Nonetheless, even if the query generation and the external trusted knowledge source are perfect and always provide a golden source, there is no guarantee that the generated responses are grounded on the source. One such factor is randomness in temperature sampling. As depicted in \citep{holtzman2019curious}, the text generated from pure sampling ($T =1.0$) is incoherent and almost unrelated to the source. The paper suggested the "unreliable tail" is to blame. The unreliable tail is composed of low probable tokens which aggregate to unignorable probability mass in sampling. When a token in the unreliable tail is sampled, there is a chance that hallucination happens. The paper proposed nucleus sampling (top-p) as a static solution to capture the region of confidence to avoid the unreliable tail.

We hypothesize that there is a trade-off between diversity and attribution in contextual predictions. This paper examines this trade-off and proposes a decoding method\footnote{Code is available at \url{https://github.com/google-research/google-research/tree/master/kl_guided_sampling}} to address it. The proposed decoder dynamically adjusts temperature at each decoding step, instead of having a fixed temperature in all decoding steps. There are two components in the mechanism: (1) A signal to indicate whether the source is relevant to the current decoding step; (2) A converter to take this signal as input and adjust sampling accordingly. For (1), we propose an original idea to leverage KL-divergence as a guiding signal to tell whether the source is relevant; For (2), we leverage the finding in \citep{aksitov2023characterizing} and take temperature as the knob for attributions, besides diversity, where the authors suggested that temperature and model size has similar scale of effect on attributions.

\section{Background}

\subsection{Contextual Predictions}

In contextual predictions, LLM is expected to generate a response that is attributable to the source, if the source is relevant. From the perspective of natural language generation (NLG) tasks, the contextual prediction task is a prefix LM task, where source is part of the prefix. For an encoder-decoder model architecture, the source is part of the input to the encoder; For a decoder-only architecture, source is part of the prefix to the decoder.

In KL-divergence guided temperature sampling, for the purpose of comparing the distributions between two parallel decodings: one with source in the input and one without, the model needs to be able to handle inputs with and without source. In other words, training data must include examples with and without sources.

For some NLG tasks, the notion of source is explicit and the model being able to handle either cases (having source or not) comes for free from the model pretraining. For a conversational QA task over QReCC datasets in Section 6.2 in \citep{aksitov2023characterizing}, the authors encapsulated conversations and an optional source into inputs with advanced promptings.

For some other NLG tasks, on the other hand, the source is implicit and the notion of not having source needs to be handled carefully. Consider a text summarization task, most training examples have pairs of a paragraph and a summary. The model has never seen empty input in training time and hence the prediction without source in inference time may be unexpected. One simple workaround is to train the model with empty inputs, as if it is a causal LM task. Please see more discussion in Section \ref{sec:summarization}.

\subsection{Temperature Sampling}

In a conventional decoding step with the temperature sampling algorithm, the temperature is used to adjust vocabulary probability mass function (PMF) according to
\begin{equation} \label{eq:temperature_sampling}
  \Pr(v_k) = \frac{e^{l_k / T}}{\sum_i e^{l_i / T}}
\end{equation}
where $v_k$ is the $k$-th vocabulary token, $l_k$ the corresponding logit, and $T$ a constant temperature. In this work, we follow the implementation of temperature sampling in the T5X library\footnote{See https://github.com/google-research/t5x.}.

In inference time, the temperature $T$ is adjusted:
\begin{itemize}
    \item When $T = 0$, the PMF becomes a Kronecker delta function, and the algorithm degenerates to a greedy algorithm, where the response is deterministic and likely repetitive;
    \item With larger $T$, the PMF becomes more evenly distributed. Tokens with low probability mass become more likely to be sampled, thus, the generated output becomes diverse but comes with a chance that hallucination happens. Randomness in sampling avoided repetitive responses in the greedy algorithm.
\end{itemize}
Please note that keeping the same temperature to all vocabulary $v_k$ helps preserve the order of the PMF, and this order helps to preserve the sensibleness of the predictions. 

\subsection{KL-Divergence}

KL-divergence is a metric to measure the statistical difference between two distributions. The formula is as follows:
\begin{equation} \label{eq:kl_divergence}
    KL(p||q) = \sum_k p_k \log \frac{p_k}{q_k}
\end{equation}
where $p = \{p_k\}$ and $q = \{q_k\}$ are two PMFs.

At each decoding step, let $p$ be the token PMF where model input contains source and $q$ the token PMF where model input contains no source. With this convention, the KL-divergence is a measure of how source matters:
\begin{itemize}
    \item If the KL-divergence is small, the distribution $p$ and $q$ are similar irregardless of whether the source is provided. In other words, the source is irrelevant;
    \item If the KL-divergence is large, the presence of the source is consequential to the PMF. In other words, the source matters.
\end{itemize}

Mathematically, we can interpret KL-divergence as the mean of the pointwise mutual information (PMI). Let $s$ be the source and $x = \{x_t\}$ the generated response. At each decoding step $t$, we have $p_k\hspace{-0.4ex}=\hspace{-0.4ex}\Pr(x_t\hspace{-0.4ex}=\hspace{-0.4ex}v_k|x_{<t},s)$ and $q_k\hspace{-0.4ex}=\hspace{-0.4ex}\Pr(x_t\hspace{-0.4ex}=\hspace{-0.4ex}v_k | x_{<t})$ where $v_k$ is the $k$-th vocabulary token, $x_{<t} = \{x_0, \dots, x_{t-1}\}$ is the set of all previous decodes, and $\Pr(\cdot)$ stands for the empirical probability calculated by the LLM with $T = 1$. As a result,
\[
  \log\frac{p_k}{q_k} = \log \frac{\Pr(x_t=v_k, s | x_{<t})}{\Pr(x_t=v_k | x_{<t})\Pr(s | x_{<t})},
\]
which defines the PMI between the source $s$ and the decode at step $t$ being $v_k$, given all previous decodes $x_{<t}$.

The PMI measures the associations between $s$ and $x_t=v_k$ given $x_{<t}$. Conditioned on previous decodes $x_{<t}$, when PMI $= 0$, they are independent; when PMI $> 0$, they are positively associated; otherwise, they are negatively associated. According to equation (\ref{eq:kl_divergence}), KL-divergence is the mean of the PMIs over the PMF $p\hspace{-0.4ex}=\hspace{-0.4ex}\Pr(x_t|x_{<t},s)$. In other words, at each decoding step before sampling, the KL-divergence tells us, on average, how this decoding step associates with the source.

\subsection{Metrics} \label{sec:metrics}

\subsubsection{Metrics for Attribution}
We follow the convention \citep{rashkin2021measuring} for the definition of attribution, and in \citep{rashkin2021measuring,honovich2022true} to cast an attribution evaluation as an instance of natural language inference (NLI) task by treating source as the premise and the response as the hypothesis.

For the conversational QA tasks, we follow the conventions in Section 5.1 of \citep{aksitov2023characterizing} and use a T5-11B model finetuned on MNLI, SNLI, FEVER, PAWS, SciTail and VitaminC \footnote{A newer version from the authors.}. We follow the flavor v3 in the paper where the NLI premise and hypothesis are "\{Source\} \{Question\}" and "\{Question\} \{Answer\}", respectively, for the QA tasks, and follow the convention to adapt the NLI when the premise is very long.

For the summarization task, we follow the conventions in Section 3 of \citep{aharoni2022mface}. We use the same model, a.k.a. a mT5-XXL model finetuned on ANLI and XNLI. We also follow the convention of having the NLI premise and hypothesis as "\{Source\}" and "\{Summary\}", respectively.

\subsubsection{Metrics for Diversity}
We adopt the likelihood evaluation metrics in \citep{holtzman2019curious}. However, we only present the experiment results with self-BLEU4. According to the paper, The metrics self-BLUE4, self-BLUE5, and zipf are similar, see Figure 8 and Section 5.2. It has been argued that perplexity is not the best metric for measuring the performance of language models because natural language does not always follow the most probable path. Finally, we dropped repeated n-gram as we don't see it as a major issue, though it exists, in our experiments.

Besides conventional metrics, we create an additional metric, var-rank, to measure the variance of the token ranks. The metric var-rank is a direct statistical measurement of how temperature affects sampling. For example, in a greedy algorithm, all tokens have rank $0$ and, hence, var-rank equals $0$. In sampling with top-k equals $40$, all token ranks are in the range of $[0, 39]$.

For each example $k$ in the evaluation dataset, we generate one decode, represented by token IDs $I_k = \{i_{kj}\}$ and the token ranks $R_k = \{r_{kj}\}$ where $i_{kj}$ is a positive integer (and not including the padding $0$s) and $r_{kj}$ is a non-negative integer. For all examples we have two lists of lists: one for token IDs $\hat{I} = \{I_k\}$ and one for token ranks $\hat{R} = \{R_k\}$. For var-rank, we flatten $\hat{I}$ into a single list and calculate the variance. For self-BLEU4, we calculate self-bleu by comparing all pairs of outputs across the entire dataset. For var-rank, the higher the better (or more diverse). For self-BLEU, the lower the better (or more diverse).

\section{KL-Divergence Guided Temperature Sampling}

\subsection{Architecture}

\begin{figure}[t]
    \centering
      \includegraphics[width=0.48\textwidth]{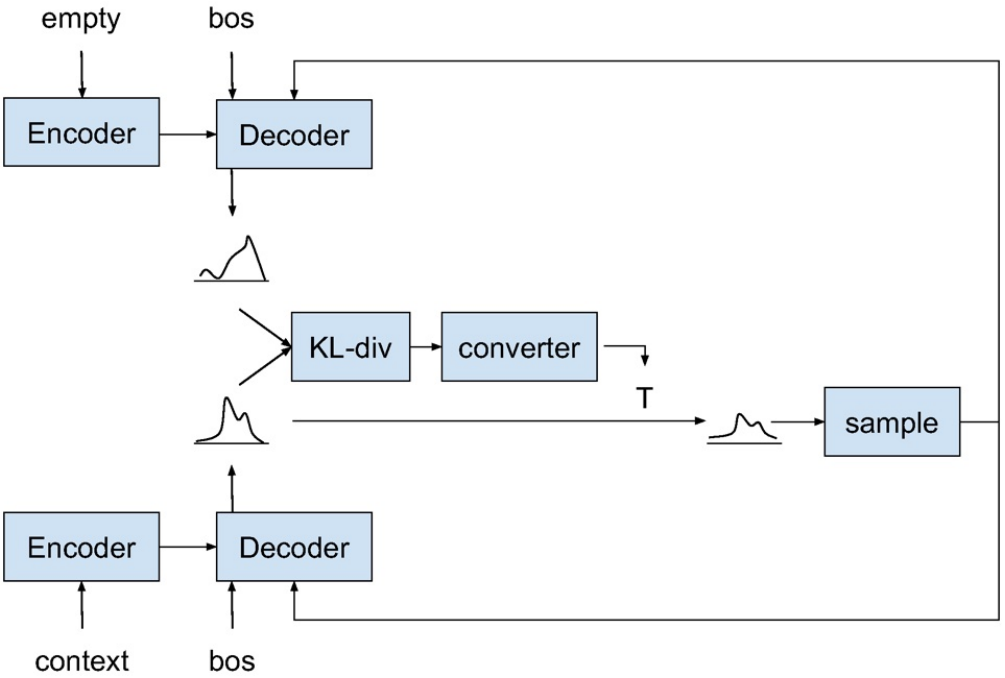}
      \caption{System architecture}
      \label{fig:system_diagram}
\end{figure}

The system architecture is illustrated in Figure \ref{fig:system_diagram}. For convenience, we take encoder-decoder models as an example, but the methodology also works for decoder-only models. Assuming data preprocessors and feature converters have prepared each example with two inputs, the model runs two parallel decodings: the bottom one with source in the input and the top one without source in the input.

In each decoding step, both decoders first compute logits and the token PMFs. The PMFs from two parallel decodings are used to calculate the KL-divergence according to the equation (\ref{eq:kl_divergence}). The converter takes this KL-divergence as input and adjusts the temperature according to the equation (\ref{eq:converter_func}), see Section \ref{sec:converter}. Finally, we apply the temperature to the logits according to the equation (\ref{eq:temperature_sampling}), followed by any additional top-k and top-p masking, and random sampling. The sampled next token is fed back to both parallel decoders for the next decoding step.

\section{Converter} \label{sec:converter}

The converter takes KL-divergence as input and computes the temperature for the decoding step. For simplicity, we use the following exponential decay function:
\begin{equation} \label{eq:converter_func}
    T = T_0 \cdot \left(\frac{1}{2}\right)^{\frac{KL(p||q)}{\sigma}}
\end{equation}
where $T_0$ is the baseline temperature and $\sigma$ is a hyperparameter to specify the half-life cycle of the decay. Please note that when $\sigma$ is very small, this function degenerates to $T = 0$; when $\sigma$ is very large, this function degenerates to $T = T_0$.

\begin{figure}[!htbp]
    \centering
      \includegraphics[width=0.48\textwidth]{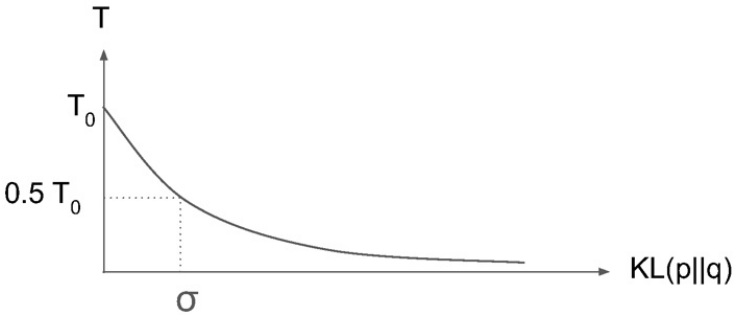}
      \caption{converter function}
      \label{fig:converter_function}
\end{figure}

\begin{figure}[!thbp]
    \centering
      \includegraphics[width=0.48\textwidth]{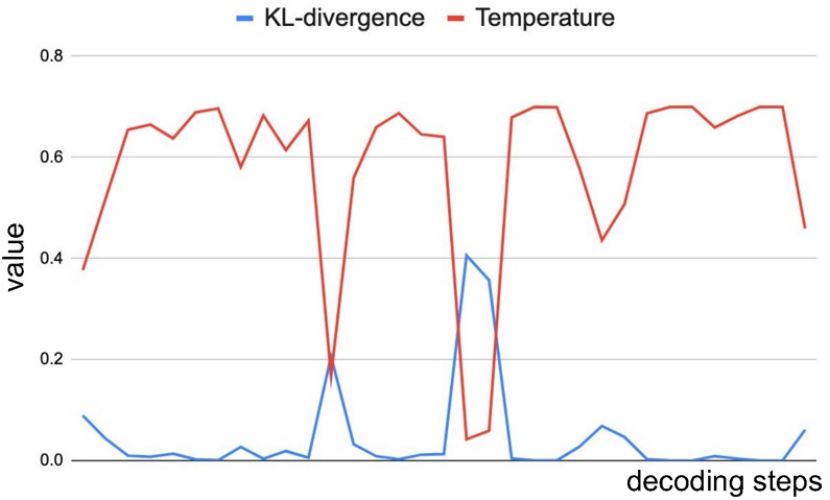}
      \caption{An example of KL-divergence to temperature sequences over decoding steps with $\sigma=0.1$.}
      \label{fig:converter_example}
\end{figure}

Figure \ref{fig:converter_example} demonstrates the KL-divergence and the corresponding temperature over decoding steps in a real example.

\section{Experiments}

\subsection{Baseline and Guided Groups} \label{sec:baseline_guided_groups}

In the following experiments, we will compare different decoding algorithms with the metrics in Section \ref{sec:metrics}. We categorize the experiments into two groups: baseline group and guided groups. In the baseline group, there are three decoding algorithms, namely \emph{baseline T} (T stands for temperature), \emph{baseline top-p}, and \emph{baseline top-k}, with the following settings:
\begin{itemize}
    \item Baseline T: top-k=$40$, top-p=$1.0$, T $\in$ $\{0$, $0.1$, $\cdots$, $0.9$, $1.0\}$
    \item Baseline top-p: top-k=all, T=$1.0$, top-p $\in$ $\{0$, $0.01$, $0.05$, $0.1$, $0.2$, $\cdots$, $0.8$, $0.9$, $0.95$, $0.99$, $1.0\}$
    \item Baseline top-k: T=$1.0$, top-p=$1.0$, top-k $\in$ $\{1$, $2$, $5$, $10$, $20$, $40$, $80$, $160$, $320$, $640$, $1280$, all$\}$
\end{itemize}
Please note that when top-p=$1.0$, it is equivalent to not applying top-p masking, and similarly for top-k=all.

\begin{figure}[!tbp]
    \centering
      \includegraphics[width=0.48\textwidth]{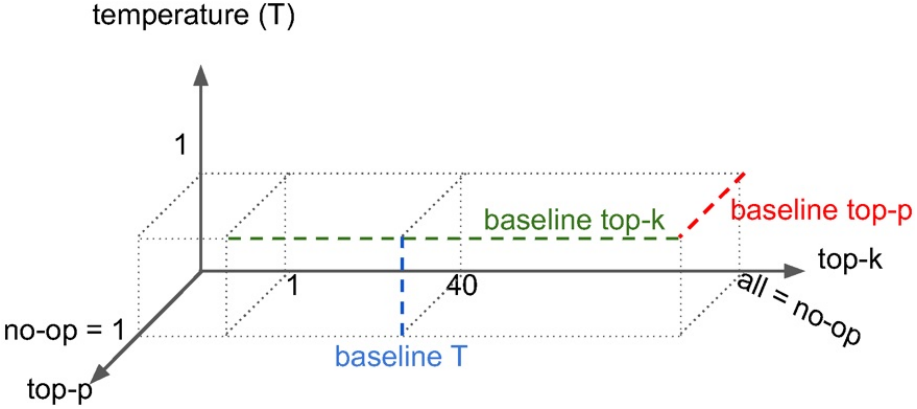}
      \caption{Baseline experiments}
      \label{fig:baseline_group}
\end{figure}

These baseline experiments are illustrated in Figure \ref{fig:baseline_group}, where the axes are T, top-k, top-p, respectively. There are three intersections between the baselines:
\begin{itemize}
    \item Baseline top-p intersects with baseline top-k at (top-k=all, top-p=$1$, T=$1$);
    \item Baseline T intersects with baseline top-k at (top-k=$40$, top-p=$1$, T=$1$);
    \item Three open-ended ends of all baselines implicitly intersect at the greedy algorithm (top-k=$1$, or top-p=$0$, or T=$0$).
\end{itemize}

In the guided groups, there are two decoding algorithms guided by KL-divergence, namely \emph{guided T} and \emph{guided top-p}, with the following settings:
\begin{itemize}
    \item Guided T: top-k=$40$, top-p=$1.0$, fix T$_0$=$0.7$ but change $\sigma \in \{1E^{-4}$, $3E^{-4}$, $1E^{-3}$, $3E^{-3}$, $1E^{-2}$, $3E^{-2}$, $0.1$, $0.3$, $1$, $3$, $\infty\}$
    \item Guided top-p: top-k=all, T$_0$=$1.0$, fix top-p=$0.95$ but change $\sigma \in \{1E^{-4}$, $3E^{-4}$, $1E^{-3}$, $3E^{-3}$, $1E^{-2}$, $3E^{-2}$, $0.1$, $0.3$, $1$, $3$, $\infty\}$
\end{itemize}
The naming may be confusing as guided top-p experiment doesn't change top-p. Instead, KL-divergence is guiding the temperature while keeping top-p fixed. We name it guided top-p as it is meant to be compared with baseline top-p. Similarly, guided T is meant to be compared with baseline T.

\begin{figure}[!tbp]
    \centering
      \includegraphics[width=0.48\textwidth]{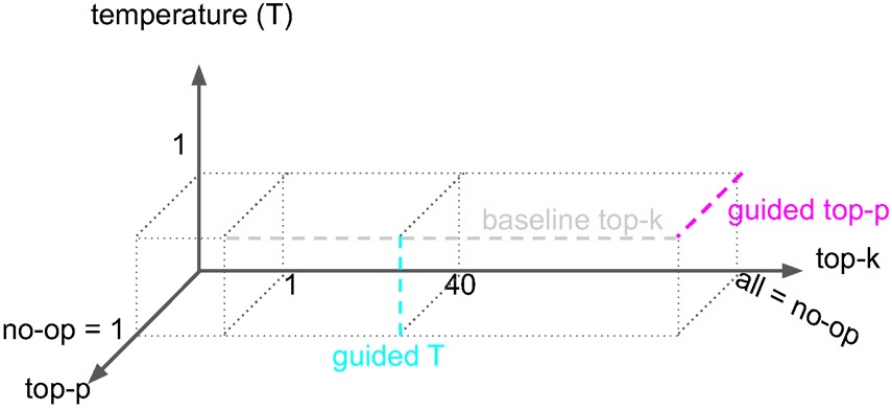}
      \caption{Guided experiments}
      \label{fig:guided_group}
\end{figure}

\section{Conversational QA tasks}

\subsection{Procedure}

We use the dataset Question Rewriting in Conversational Context (QReCC) \citep{anantha2020open} – a knowledge intensive open-domain QA dataset consisting of 14K conversations with 81K QA pairs. Each conversation consists of a series of questions and answers as conversation history, followed by a final query. A golden answer, along with the webpage the answer was extracted from, is also provided for each conversation. 

We follow the same setting in \citep{aksitov2023characterizing} to preprocess and filter datasets, and construct queries to a LLM. In preprocessings, we use a fully decontextualized version of QReCC in our experiments. The webpage where the answer was extracted from is crawled and processed as text. We follow Table 3 in the paper to apply filters to the dataset to remove examples that are not well-formed. We take the final 2829 examples in the training split for our experiments.

We format the input queries to PaLM \citep{chowdhery2022palm} models with advanced promptings as described in Section 6.2 in \citep{aksitov2023characterizing}:
\begin{itemize}
    \item For the input with source: without instructions, with the golden evidence, and with dialog history;
    \item For the model input without instructions: without instructions and evidence, and with dialog history.
\end{itemize}
We take the first turn (ends by the first [eot]) in the generated response as the model's response to the conversation.

Figure \ref{fig:qrecc_palm8b} shows the attribution-diversity trade-off curves of different decoding algorithms over the PaLM-8B model. Each datapoint is the summary of attribution and diversity of all 2829 QReCC examples under a specific algorithm. For each example, we construct and send the input query to the PaLM-8B model to generate a response and token ranks in all decoding steps. We grade attribution with E2E NLI on each response and take the mean over all responses (y-axis in Figure \ref{fig:qrecc_palm8b}). We grade diversity by var-rank and self-BLEU4. For the former, we flatten all token ranks from all examples and calculate the variance (x-axis in Figure \ref{fig:qrecc_palm8b} left); For the latter, we follow the convention in \citep{holtzman2019curious} (x-axis in Figure \ref{fig:qrecc_palm8b} right). Please note that the x-axis is presented in log scale for var-rank. Since the greedy algorithm has rank to be all 0s, the variance of the rank (x-axis) is at negative infinity, causing the artifact of the horizontal lines in Figure \ref{fig:qrecc_palm8b} left. 

\subsection{Results}

\begin{figure*}[!tbp]
    \centering
      \includegraphics[width=0.9\textwidth]{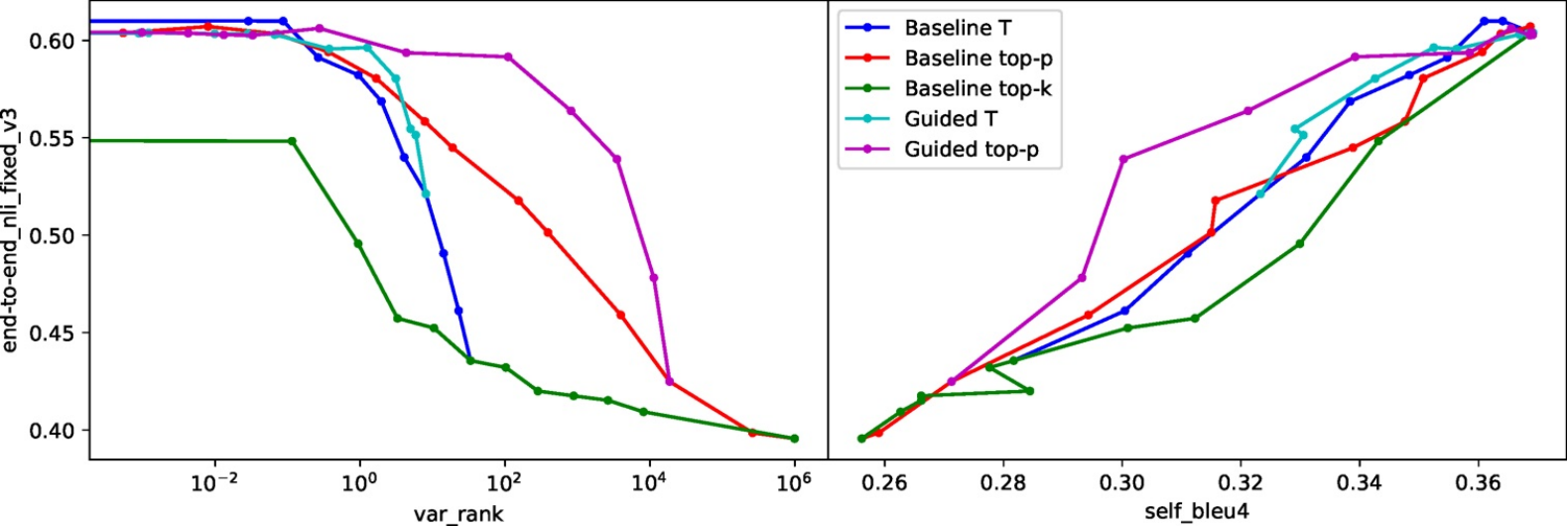}
      \caption{QReCC dataset over PaLM-8B model}
      \label{fig:qrecc_palm8b}
\end{figure*}

First, let's look at three baseline experiments in Figure \ref{fig:qrecc_palm8b} left.
As sanity checks, baseline top-k (green) intersects baseline T (blue) and baseline top-p (red) once for each, and they intersect at x=$-\infty$ (greedy algorithm) implicitly. Furthermore, all three lines show the trade-off between diversity and attributions. Ideally, we prefer to have a decoding algorithm with its datapoint at the top-right corner, meaning that the algorithm has high diversity and high attributions. From this perspective, baseline top-p is better than baseline T, and baseline top-k is the worst.

To compare the baseline group with the guided group, we compare baseline T (blue) with guided T (cyan), and baseline top-p (red) with guided top-p (magenta), respectively. Evidently, guided T outperforms baseline T marginally while guided top-p outperforms baseline top-p significantly. This is also justified by Figure \ref{fig:qrecc_palm8b} right, where the top-left corner is better.

Concrete examples of the generated texts are in Appendix \ref{sec:appendix_qrecc}. We present several examples to illustrate that the algorithm works as intended. For illustration, we select one algorithm from the guided top-p and two algorithms from baseline top-p so that they are aligned horizontally or vertically for comparison. We also color text tokens by temperatures. These examples are cherry-picked. In practice, many examples in the guided top-p are err on either side of the baseline counterparts. That is, there are many examples with all responses being identical (or not diverse), and many with responses hallucinating.

\begin{figure*}[!tbp]
    \centering
      \includegraphics[width=0.9\textwidth]{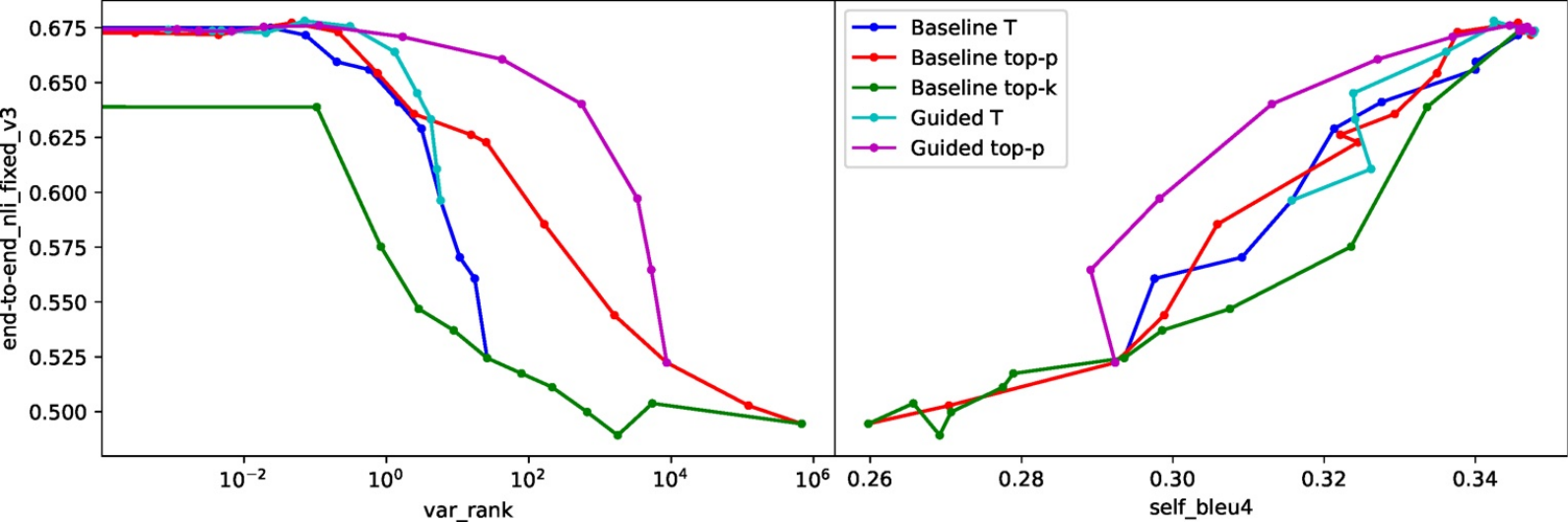}
      \caption{QReCC dataset over PaLM-62B model}
      \label{fig:qrecc_palm62b}
\end{figure*}

We applied the same set of experiments to the PaLM-62B model in Figure \ref{fig:qrecc_palm62b}. It is expected that PaLM-62B is more performant than PaLM-8B, and this is justified by the fact that all curves shift upward. For example, with the greedy algorithm, attribution is increased from 61\% (PaLM-8B) to 67.5\% (PaLM-62B). For the comparison of the trade-off curves among baseline and guided groups, the trend is similar:
\begin{itemize}
    \item Comparing the baseline experiments: baseline top-p $>$ baseline T $>$ baseline top-k;
    \item Comparing the baseline and guided experiments, guided top-p $\gg$ baseline top-p and guided T $>$ baseline T.
\end{itemize}

In summary, conversational QA is a perfect example to illustrate the advantage of the KL-divergence guided temperature sampling. In a conversational QA setting, there are several previous turns to act as few-shots to the final QA task. One nature of QA tasks is that the information provided by the source is concentrated in a few response tokens, which makes room to selectively change temperatures in decoding steps. Although the proposed approach cannot mitigate the hallucinations in low temperatures and greedy algorithms, these errors can be mitigated through increasing the model size.

\begin{figure*}[!tbp]
    \centering
      \includegraphics[width=0.9\textwidth]{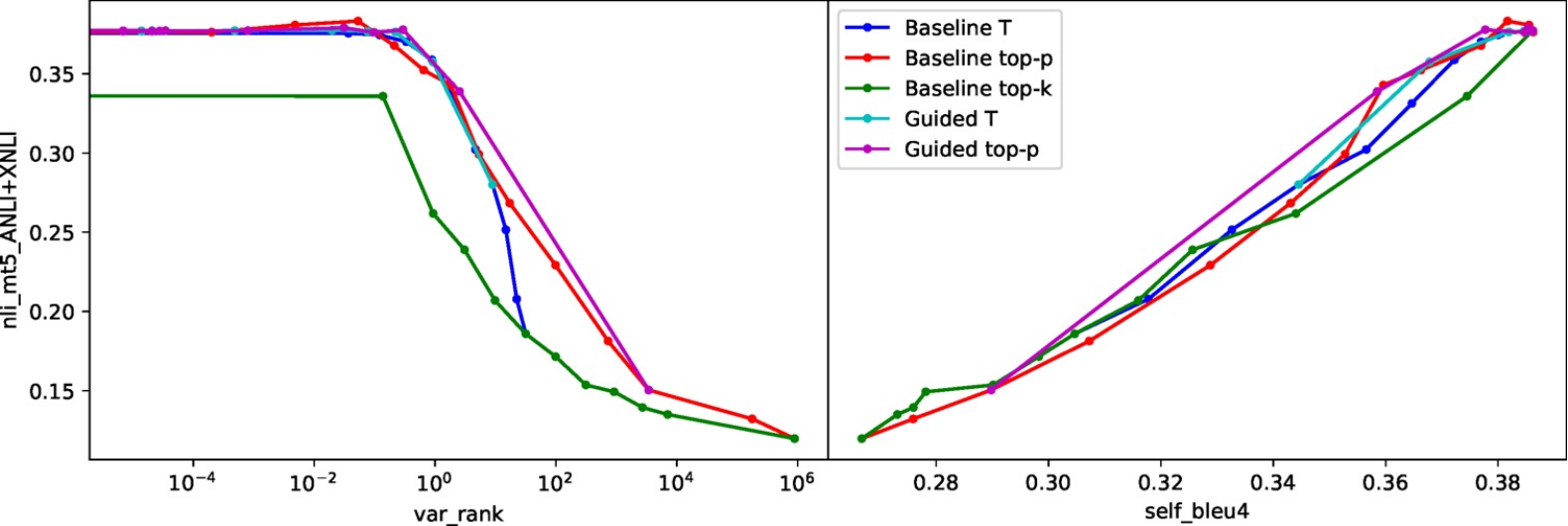}
      \caption{XLSum dataset over MT5 XL model}
      \label{fig:xlsum_mt5xl_mandatory}
\end{figure*}

\begin{figure*}[!tbp]
    \centering
      \includegraphics[width=0.9\textwidth]{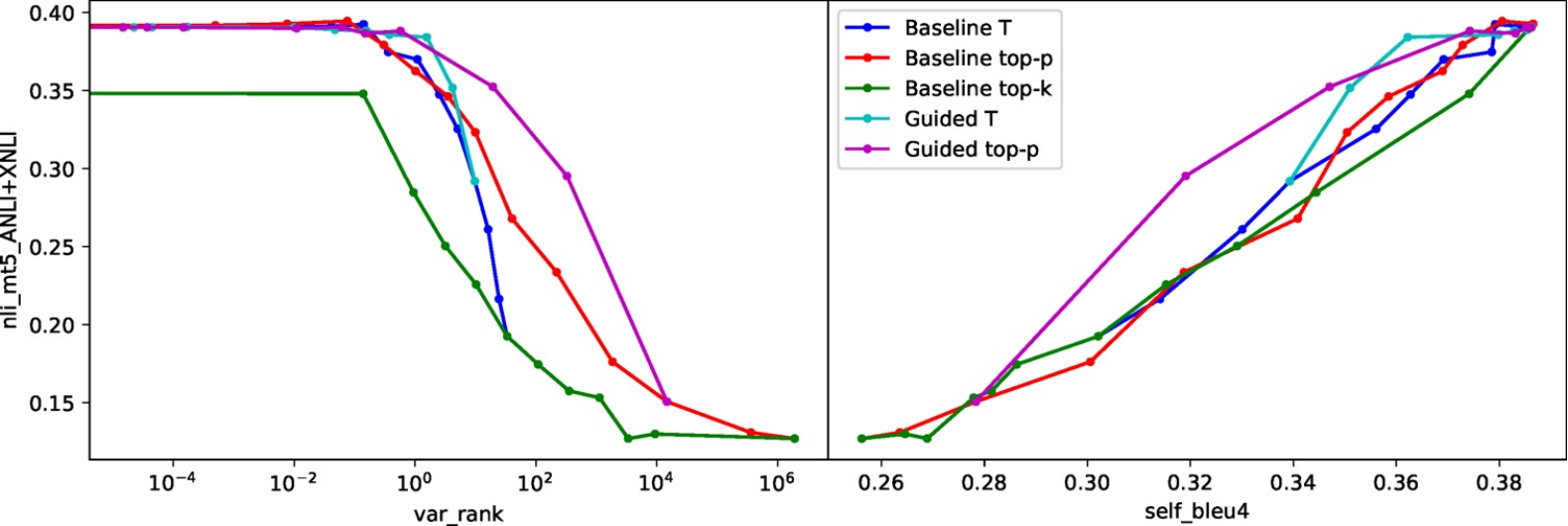}
      \caption{XLSum dataset over MT5 XL model with optional inputs. In all figures, y-axis represents attribution with the automatic metric E2E NLI, and x-axis represents diversity with the automatic metrics var-rank (left) and self-BLEU4 (right). Each figure compares the trade-off curves of $5$ experiments. In particular, we compare baseline T with guided T, and baseline top-p with guided top-p.}
      \label{fig:xlsum_mt5xl_empty}
\end{figure*}

\section{Summarization tasks} \label{sec:summarization}

\subsection{Procedure}

We take XLSum \citep{hasan2021xl}, a multilingual abstractive summarization for 44 Languages, as the finetuning and evaluation datasets. We first finetune the pretrained MT5-XL model \citep{xue2020mt5} with the XLSum training split for $15000$ steps, batch\_size $16$, sequence length = \{'inputs': $1024$, 'targets': $128$\}, and with all $44$ languages. We evaluate this finetuned model over the validation split for $3000$ examples in $13$ languages supported by XNLI (for valid attribution metric).

However, the notion of not having a source is not well-posed. Conventionally, the input to the model is the article to be summarized. While it is natural to define the whole input as the source, implying that not having source refers to an empty input, the LLM is never finetuned with empty inputs in the summarization tasks. 

\subsection{Results}

To begin with, let's naively treat all input text as a source. The results are shown in Figure \ref{fig:xlsum_mt5xl_mandatory}. The guided group is at least as performant as that of its corresponding baseline group. The datapoints in the guided group are shifted to the left, meaning that the majority of the KL-divergences of the guided group are in [$3.0$, $\infty$). This is because the model has never trained with empty inputs, as a result, the PMF with empty inputs becomes very different from that with normal inputs, which in turns causes the KL-divergence to be always large.

To overcome the issue, we modified the finetune summarization tasks to include both intact and empty inputs, so that the model will summarize the input article when it exists; otherwise, it will generate/hallucinate sensible short summary texts. In Figure \ref{fig:xlsum_mt5xl_empty} left, we notice that the guided top-p (magenta) is more performant than the guided top-p (red), and similarly for the comparison between guided T (cyan) and baseline T (blue), although the gap is smaller. In other words, for the same diversity, guided top-p has higher attribution than baseline top-p; for the same attribution, guided top-p has higher diversity than baseline top-p. A similar conclusion can be drawn from Figure \ref{fig:xlsum_mt5xl_empty} right, where the diversity axis (x-axis) is replaced by self-BLEU4. Concrete examples of the generated texts to compare baseline top-p (top-p=$0.4$ and $0.7$) and guided top-p ($\sigma$=$1.0$) are in Appendix \ref{sec:appendix_xlsum}.

\section{Discussion} \label{sec:discussion}

According to the experiments, we justified the hypothesis that there is a trade-off between attribution and diversity, and our proposed algorithm mitigates the trade-off. However, it is evident that the gain in conversational QA tasks is more significant than that in the summarization tasks.

For many NLG tasks, the source only matters for a few tokens in the response, and the rest of the response tokens are just language that glues the information together. For example, in a QA task with the question "What is the weather today?" along with source from a weather forecast website, the response could be:
\begin{itemize}
    \item The weather today is  \underline{rainy}.
    \item It is \underline{rainy} today.
    \item The answer is \underline{rainy}.
\end{itemize}
Notice that only the token \underline{rainy} is relevant to the source. Because the relevance is only concentrated to a few tokens only, this provides the opportunity to selectively adjust temperatures.

The methodology is not perfect. One caveat is that the computation is doubled as we are having two parallel decodings. Another caveat is that when the KL-divergence is small, the source can be truly irrelevant, or the source is relevant but not consequential, \emph{e.g.} the model already knows the fact from its memory, so the source doesn't provide additional information. Having a low KL-divergence does not necessarily mean the source is irrelevant, but having a high KL-divergence definitely means the source matters.

\section{Conclusion}
This paper proposed a decoding algorithm to improve LLM attributions when source is present. The algorithm consists of two parallel decoding steps, one with source in the input and one without. The KL-divergence between two token distributions indicates how the source is relevant to the token to be sampled in the decoding step. We use this signal to adjust the sampling temperature to improve attribution when source is relevant, and to improve diversity when source is irrelevant. Although the algorithm requires doubling the computations for parallel decodings, our experiments on conversational and summarization tasks show improvements over the conventional top-k and top-p algorithms, in particular, with large margin over conversational QA tasks.

\section*{Limitations}
Compared with the conventional temperature sampling, our proposed algorithms requires twice computations for two parallel decodings. State of the art LLMs ($\sim$ O(100B) parameters) are expensive to serve. Any algorithm that further increases the computation may hinder its real world applicability.

On the other hand, the algorithm also requires models to be large enough for two reasons: First, the attribution of the algorithm is roughly upper bounded by the greedy algorithm of the LLM. The greedy algorithm's performance is improved as the model size increases. Next, the algorithm relies on the in-context learning in the LLM it is paired with. In particular, the two token PMFs, one with source in the input and one without, should be selectively different according to its relevance to the source. The in-context learning is an emergent ability that only appears after a model is large enough.

\section*{Ethics Statement}
The proposed sampling algorithm requires to be paired with an additional LLM and source providers, e.g. retrieval systems or search engines. The integrated system may inherit the bias and privacy concerns in the selected LLMs and source providers. Compared with temperature sampling, the responses generated by our proposed algorithm are within the counter of those of the temperature sampling. Our proposed algorithm only eliminates (and does not add) responses that are likely hallucinating, when compared with its baseline.

\bibliography{anthology,custom}
\bibliographystyle{acl_natbib}

\appendix \label{sec:appendix}

\section{Examples from QReCC}  \label{sec:appendix_qrecc}

According to Figure \ref{fig:qrecc_palm8b}, we choose guided top-p ($\sigma$=$0.1$), baseline top-p (=$0.3$), and baseline top-p (=$0.7$) for comparison. Responses are deduplicated and some source text is skipped to save space. For each example, we decode the same input $10$ times. See Table \ref{tab:example_conversational_qa_1}, Table \ref{tab:example_conversational_qa_2}, and Table \ref{tab:example_conversational_qa_3}.

\input{example_conversational_qa.tex}

\section{Examples from XLSum} \label{sec:appendix_xlsum}

According to Figure \ref{fig:xlsum_mt5xl_empty}, we choose algorithms guided top-p ($\sigma$=$1.0$), baseline top-p (=$0.4$), and baseline top-p (=$0.7$) for comparison. For each example, we decode the same input $5$ times. See Table \ref{tab:example_summarization_1}, Table \ref{tab:example_summarization_2}, and Table \ref{tab:example_summarization_3}.

\input{example_summarization.tex}

\end{document}

%% file: example_conversational_qa.tex
\begin{table*}
\centering
\begin{tabular}{p{0.97\linewidth}}
\hline
\textbf{Query} \\
Fact: Bure was selected 113th overall in the sixth round Draft by the Vancouver Canucks, following his rookie season with CSKA Moscow. The pick was controversial, as the Canucks had chosen him seemingly a year ahead of his eligible draft season. At the age of 18, he was available to be chosen in the first three rounds of the draft, but to be selected any later, he would have needed to play at least two seasons—with a minimum of 11 games per season—for his elite-level Soviet club, the Central Red Army. [22] While most teams believed he was ineligible, the Canucks' head scout at the time, Mike Penny, discovered Bure had played in additional exhibition and international games to make him an eligible late-round draft choice a year early. [23] [24] Jack Button, the Washington Capitals ' director of player personnel, admitted "everybody would have taken him earlier. We assumed he was not eligible... you've got to give the Canucks credit for doing their homework." [25] Several other teams either had similar knowledge or had pursued Bure, but there was confusion as to the legitimacy of the extra games. The Detroit Red Wings had asked league vice president Gil Stein about Bure's availability before their fifth-round pick, but were told he was not eligible. [22] They later decided to select him with their sixth-round pick, 116th overall, and settle his eligibility later. The Canucks selected Bure three picks ahead of Detroit's turn. [26] Meanwhile, Winnipeg Jets general manager Mike Smith , claimed he made an offer to the Soviet Ice Hockey Federation that would involve three years of transfer payments before Bure would be allowed to join the Jets; however Smith did not have any plans to draft Bure in 1989 as he believed he was ineligible. [27] 
\newline \newline
[eot] 0 -1 0 What country was Pavel Bure born? [eot] 1 0 1 Pavel Bure was born in Moscow in 1971. [eot] 2 1 0 What team did Pavel Bure play for as an adult? [eot] 3 2 1  \newline
\\
\textbf{Greedy algorithm} \\
+ \hlc[pink!0]{ The}\hlc[pink!0]{ Vancouver}\hlc[pink!0]{ Canucks}\hlc[pink!0]{.}\hlc[pink!0]{ } [x10]
\\
\textbf{Baseline top-p (=0.3)} \\
+ \hlc[pink!100]{ Pavel}\hlc[pink!100]{ Bure}\hlc[pink!100]{ played}\hlc[pink!100]{ for}\hlc[pink!100]{ the}\hlc[pink!100]{ Vancouver}\hlc[pink!100]{ Canucks}\hlc[pink!100]{.}\hlc[pink!100]{ } [x4] \newline
+ \hlc[pink!100]{ The}\hlc[pink!100]{ Vancouver}\hlc[pink!100]{ Canucks}\hlc[pink!100]{.}\hlc[pink!100]{ } [x6]
\\
\textbf{Baseline top-p (=0.7)} \\
+ \hlc[pink!100]{ Pavel}\hlc[pink!100]{ Bure}\hlc[pink!100]{ played}\hlc[pink!100]{ for}\hlc[pink!100]{ the}\hlc[pink!100]{ Vancouver}\hlc[pink!100]{ Canucks}\hlc[pink!100]{ from}\hlc[pink!100]{ }\hlc[pink!100]{1}\hlc[pink!100]{9}\hlc[pink!100]{9}\hlc[pink!100]{1}\hlc[pink!100]{ to}\hlc[pink!100]{ }\hlc[pink!100]{1}\hlc[pink!100]{9}\hlc[pink!100]{9}\hlc[pink!100]{4}\hlc[pink!100]{.}\hlc[pink!100]{ } \newline
+ \hlc[pink!100]{ Vancouver}\hlc[pink!100]{ Canucks}\hlc[pink!100]{.}\hlc[pink!100]{ } [x3] \newline
+ \hlc[pink!100]{ Pavel}\hlc[pink!100]{ Bure}\hlc[pink!100]{ played}\hlc[pink!100]{ for}\hlc[pink!100]{ the}\hlc[pink!100]{ Vancouver}\hlc[pink!100]{ Canucks}\hlc[pink!100]{ in}\hlc[pink!100]{ }\hlc[pink!100]{1}\hlc[pink!100]{9}\hlc[pink!100]{9}\hlc[pink!100]{1}\hlc[pink!100]{-}\hlc[pink!100]{9}\hlc[pink!100]{2}\hlc[pink!100]{,}\hlc[pink!100]{ and}\hlc[pink!100]{ }\hlc[pink!100]{1}\hlc[pink!100]{9}\hlc[pink!100]{9}\hlc[pink!100]{2}\hlc[pink!100]{-}\hlc[pink!100]{9}\hlc[pink!100]{3}\hlc[pink!100]{.}\hlc[pink!100]{<0x0A>}\hlc[pink!100]{<0x0A>}\hlc[pink!100]{Source}\hlc[pink!100]{:}\hlc[pink!100]{ https}\hlc[pink!100]{://}\hlc[pink!100]{en}\hlc[pink!100]{.}\hlc[pink!100]{wikipedia}\hlc[pink!100]{.}\hlc[pink!100]{org}\hlc[pink!100]{/}\hlc[pink!100]{wiki}\hlc[pink!100]{/}\hlc[pink!100]{P}\hlc[pink!100]{avel}\hlc[pink!100]{\_}\hlc[pink!100]{Bu}\hlc[pink!100]{re}\hlc[pink!100]{ } \newline
+ \hlc[pink!100]{ Bure}\hlc[pink!100]{ played}\hlc[pink!100]{ for}\hlc[pink!100]{ the}\hlc[pink!100]{ Washington}\hlc[pink!100]{ Capitals}\hlc[pink!100]{ from}\hlc[pink!100]{ }\hlc[pink!100]{1}\hlc[pink!100]{9}\hlc[pink!100]{9}\hlc[pink!100]{3}\hlc[pink!100]{ to}\hlc[pink!100]{ }\hlc[pink!100]{1}\hlc[pink!100]{9}\hlc[pink!100]{9}\hlc[pink!100]{7}\hlc[pink!100]{.}\hlc[pink!100]{ } \newline
+ \hlc[pink!100]{ The}\hlc[pink!100]{ Florida}\hlc[pink!100]{ Panthers}\hlc[pink!100]{.}\hlc[pink!100]{ } [x2] \newline
+ \hlc[pink!100]{ Pavel}\hlc[pink!100]{ Bure}\hlc[pink!100]{ played}\hlc[pink!100]{ for}\hlc[pink!100]{ the}\hlc[pink!100]{ Vancouver}\hlc[pink!100]{ Canucks}\hlc[pink!100]{ for}\hlc[pink!100]{ }\hlc[pink!100]{1}\hlc[pink!100]{4}\hlc[pink!100]{ years}\hlc[pink!100]{,}\hlc[pink!100]{ playing}\hlc[pink!100]{ from}\hlc[pink!100]{ }\hlc[pink!100]{1}\hlc[pink!100]{9}\hlc[pink!100]{9}\hlc[pink!100]{1}\hlc[pink!100]{ to}\hlc[pink!100]{ }\hlc[pink!100]{2}\hlc[pink!100]{0}\hlc[pink!100]{0}\hlc[pink!100]{4}\hlc[pink!100]{.}\hlc[pink!100]{ } \newline
+ \hlc[pink!100]{ The}\hlc[pink!100]{ New}\hlc[pink!100]{ York}\hlc[pink!100]{ Rangers}\hlc[pink!100]{.}\hlc[pink!100]{ }
\\
\textbf{Guided top-p (=0.1)} \\
+ \hlc[pink!54]{ The}\hlc[pink!6]{ Vancouver}\hlc[pink!98]{ Canucks}\hlc[pink!90]{.}\hlc[pink!93]{ } \newline
+ \hlc[pink!52]{ Pavel}\hlc[pink!99]{ Bure}\hlc[pink!94]{ played}\hlc[pink!97]{ in}\hlc[pink!81]{ the}\hlc[pink!72]{ National}\hlc[pink!99]{ Hockey}\hlc[pink!99]{ League}\hlc[pink!94]{ for}\hlc[pink!89]{ the}\hlc[pink!41]{ Vancouver}\hlc[pink!99]{ Canucks}\hlc[pink!94]{.}\hlc[pink!95]{ } \newline
+ \hlc[pink!52]{ Pavel}\hlc[pink!99]{ Bure}\hlc[pink!94]{ played}\hlc[pink!97]{ professional}\hlc[pink!66]{ hockey}\hlc[pink!83]{ with}\hlc[pink!50]{ the}\hlc[pink!44]{ Vancouver}\hlc[pink!99]{ Canucks}\hlc[pink!56]{ of}\hlc[pink!99]{ the}\hlc[pink!47]{ National}\hlc[pink!99]{ Hockey}\hlc[pink!98]{ League}\hlc[pink!91]{.} \newline
+ \hlc[pink!54]{ Pavel}\hlc[pink!99]{ Bure}\hlc[pink!93]{ played}\hlc[pink!97]{ for}\hlc[pink!84]{ the}\hlc[pink!17]{ Vancouver}\hlc[pink!99]{ Canucks}\hlc[pink!83]{.}\hlc[pink!93]{ } [x4] \newline
+ \hlc[pink!54]{ Pavel}\hlc[pink!99]{ Bure}\hlc[pink!93]{ played}\hlc[pink!97]{ for}\hlc[pink!84]{ the}\hlc[pink!17]{ Vancouver}\hlc[pink!99]{ Canucks}\hlc[pink!83]{ for}\hlc[pink!66]{ }\hlc[pink!96]{2}\hlc[pink!89]{0}\hlc[pink!80]{ seasons}\hlc[pink!95]{.}\hlc[pink!88]{ } \newline
+ \hlc[pink!54]{ Pavel}\hlc[pink!99]{ Bure}\hlc[pink!93]{ played}\hlc[pink!97]{ }\hlc[pink!83]{2}\hlc[pink!94]{0}\hlc[pink!87]{ NHL}\hlc[pink!88]{ seasons}\hlc[pink!90]{ for}\hlc[pink!90]{ the}\hlc[pink!53]{ Vancouver}\hlc[pink!99]{ Canucks}\hlc[pink!94]{ (}\hlc[pink!83]{1}\hlc[pink!98]{9}\hlc[pink!50]{8}\hlc[pink!94]{9}\hlc[pink!1]{-}\hlc[pink!86]{1}\hlc[pink!99]{9}\hlc[pink!99]{9}\hlc[pink!83]{6}\hlc[pink!84]{)}\hlc[pink!91]{ and}\hlc[pink!69]{ Florida}\hlc[pink!99]{ Panthers}\hlc[pink!99]{ (}\hlc[pink!99]{1}\hlc[pink!99]{9}\hlc[pink!99]{9}\hlc[pink!98]{6}\hlc[pink!84]{-}\hlc[pink!97]{2}\hlc[pink!99]{0}\hlc[pink!99]{0}\hlc[pink!90]{2}\hlc[pink!88]{)}\hlc[pink!75]{ } \newline
+ \hlc[pink!54]{ Pavel}\hlc[pink!99]{ Bure}\hlc[pink!93]{ played}\hlc[pink!97]{ professional}\hlc[pink!68]{ hockey}\hlc[pink!85]{ in}\hlc[pink!87]{ the}\hlc[pink!51]{ National}\hlc[pink!99]{ Hockey}\hlc[pink!99]{ League}\hlc[pink!89]{ (}\hlc[pink!94]{NHL}\hlc[pink!96]{)}\hlc[pink!96]{ for}\hlc[pink!93]{ the}\hlc[pink!58]{ Vancouver}\hlc[pink!99]{ Canucks}\hlc[pink!73]{,}\hlc[pink!28]{ New}\hlc[pink!89]{ York}\hlc[pink!67]{ Rangers}\hlc[pink!97]{ and}\hlc[pink!76]{ Florida}\hlc[pink!99]{ Panthers}\hlc[pink!94]{.}\hlc[pink!87]{<0x0A>}\hlc[pink!97]{<0x0A>}\hlc[pink!73]{He}\hlc[pink!90]{ scored}\hlc[pink!96]{ }\hlc[pink!92]{3}\hlc[pink!95]{4}\hlc[pink!94]{6}\hlc[pink!98]{ goals}\hlc[pink!99]{ and}\hlc[pink!91]{ }\hlc[pink!98]{7}\hlc[pink!99]{2}\hlc[pink!99]{1}\hlc[pink!91]{ assists}\hlc[pink!97]{ for}\hlc[pink!97]{ }\hlc[pink!99]{1}\hlc[pink!98]{,}\hlc[pink!99]{0}\hlc[pink!99]{6}\hlc[pink!99]{7}\hlc[pink!99]{ points}\hlc[pink!98]{,}\hlc[pink!88]{ as}\hlc[pink!99]{ well}\hlc[pink!99]{ as}\hlc[pink!96]{ }\hlc[pink!99]{1}\hlc[pink!94]{,}\hlc[pink!99]{4}\hlc[pink!99]{3}\hlc[pink!99]{7}\hlc[pink!99]{ penalty}\hlc[pink!99]{ minutes}\hlc[pink!96]{ in}\hlc[pink!99]{ }\hlc[pink!92]{1}\hlc[pink!99]{,}\hlc[pink!95]{3}\hlc[pink!99]{2}\hlc[pink!99]{3}\hlc[pink!97]{ career}\hlc[pink!94]{ games}\hlc[pink!90]{.}\hlc[pink!67]{ }
\\
\hline
\end{tabular}
\caption{Prediction examples of the QReCC conversational QA task.}
\label{tab:example_conversational_qa_1}
\end{table*}

\begin{table*}
\centering
\begin{tabular}{p{0.97\linewidth}}
\hline
\textbf{Query} \\
Fact: 4.5.4 Ecuadorian 4.5.5 Colombian 4.5.6 Salvadoran 5 See also 6 References 7 External links Population [ edit ] Historical population Year Pop. ±\% 1698 4,937 — 1712 5,840 +18.3\% 1723 7,248 +24.1\% 1737 10,664 +47.1\% 1746 11,717 +9.9\% 1756 13,046 +11.3\% 1771 21,863 +67.6\% 1790 33,131 +51.5\% 1800 60,515 +82.7\% 1810 96,373 +59.3\% 1820 123,706 +28.4\% 1830 202,589 +63.8\% 1840 312,710 +54.4\% 1850 515,547 +64.9\% 1860 813,669 +57.8\% 1870 942,292 +15.8\% 1880 1,206,299 +28.0\% 1890 1,515,301 +25.6\% 1900 3,437,202 +126.8\% 1910 4,766,883 +38.7\% 1920 5,620,048 +17.9\% 1930 6,930,446 +23.3\% 1940 7,454,995 +7.6\% 1950 7,891,957 +5.9\% 1960 7,781,984 -1.4\% 1970 7,894,862 +1.5\% 1980 7,071,639 -10.4\% 1990 7,322,564 +3.5\% 2000 8,008,288 +9.4\% 2010 8,175,133 +2.1\% 2014 8,491,079 +3.9\% 2018 8,398,748 [19] -1.1\% 1880 \& 1890 figures include part of the Bronx. Beginning with 1900, figures are for consolidated city of five boroughs. For the same area before 1900, see \#Historical Population data , below. Sources: 1698–1771, [15] 1790–1990, [16] 2000 and 2010 Censuses, [17] and 2014 Census estimates [18] New York City is the most populous city in the United States, with an estimated 8,491,079 people living in the city, according to U.S. Census estimates dating July 1, 2014 [18] (up from 8,175,133 in 2010; 8.0 million in 2000; and 7.3 million in 1990). [17] This amounts to about 40\% of New York State's population and a similar percentage of the metropolitan regional population. New York's two key demographic features are its population density and cultural diversity . The city's population density of 26,403 people per square mile (10,194/km²), makes it the densest of any American municipality with a population above 100,000. [20] Manhattan 's population density is 66,940 people per square mile (25,846/km²), highest of any county in the United States. [21] [22] \vspace{-14pt}
\newline \newline 
[eot] 0 -1 0 what percent of the us population lives in new york city? [eot] 1 0 1 2.6\% of the US population lives in New York City [eot] 2 1 0 how many people live in New York City? [eot] 3 2 1
\\
\textbf{Greedy algorithm} \\
+ \hlc[pink!0]{ }\hlc[pink!0]{8}\hlc[pink!0]{.}\hlc[pink!0]{4}\hlc[pink!0]{9}\hlc[pink!0]{1}\hlc[pink!0]{ million}\hlc[pink!0]{ people}\hlc[pink!0]{ live}\hlc[pink!0]{ in}\hlc[pink!0]{ New}\hlc[pink!0]{ York}\hlc[pink!0]{ City}\hlc[pink!0]{ } [x10]
\\
\textbf{Baseline top-p (=0.3)} \\
+ \hlc[pink!100]{ }\hlc[pink!100]{8}\hlc[pink!100]{.}\hlc[pink!100]{4}\hlc[pink!100]{9}\hlc[pink!100]{1}\hlc[pink!100]{ million}\hlc[pink!100]{ people}\hlc[pink!100]{ live}\hlc[pink!100]{ in}\hlc[pink!100]{ New}\hlc[pink!100]{ York}\hlc[pink!100]{ City}\hlc[pink!100]{ } [x10]
\\
\textbf{Baseline top-p (=0.7)} \\
+ \hlc[pink!100]{ }\hlc[pink!100]{8}\hlc[pink!100]{.}\hlc[pink!100]{4}\hlc[pink!100]{9}\hlc[pink!100]{1}\hlc[pink!100]{ million}\hlc[pink!100]{ people}\hlc[pink!100]{ live}\hlc[pink!100]{ in}\hlc[pink!100]{ New}\hlc[pink!100]{ York}\hlc[pink!100]{ City}\hlc[pink!100]{ } [x2] \newline
+ \hlc[pink!100]{ }\hlc[pink!100]{8}\hlc[pink!100]{.}\hlc[pink!100]{4}\hlc[pink!100]{9}\hlc[pink!100]{ million}\hlc[pink!100]{ people}\hlc[pink!100]{ live}\hlc[pink!100]{ in}\hlc[pink!100]{ New}\hlc[pink!100]{ York}\hlc[pink!100]{ City}\hlc[pink!100]{.}\hlc[pink!100]{ } \newline
+ \hlc[pink!100]{ }\hlc[pink!100]{8}\hlc[pink!100]{,}\hlc[pink!100]{4}\hlc[pink!100]{9}\hlc[pink!100]{1}\hlc[pink!100]{,}\hlc[pink!100]{0}\hlc[pink!100]{7}\hlc[pink!100]{9}\hlc[pink!100]{ people}\hlc[pink!100]{ live}\hlc[pink!100]{ in}\hlc[pink!100]{ New}\hlc[pink!100]{ York}\hlc[pink!100]{ City}\hlc[pink!100]{ } \newline
+ \hlc[pink!100]{ more}\hlc[pink!100]{ than}\hlc[pink!100]{ }\hlc[pink!100]{8}\hlc[pink!100]{.}\hlc[pink!100]{5}\hlc[pink!100]{ million}\hlc[pink!100]{ people}\hlc[pink!100]{ } \newline
+ \hlc[pink!100]{ more}\hlc[pink!100]{ than}\hlc[pink!100]{ the}\hlc[pink!100]{ population}\hlc[pink!100]{ of}\hlc[pink!100]{ any}\hlc[pink!100]{ other}\hlc[pink!100]{ state}\hlc[pink!100]{.}\hlc[pink!100]{ } \newline
+ \hlc[pink!100]{ }\hlc[pink!100]{8}\hlc[pink!100]{,}\hlc[pink!100]{1}\hlc[pink!100]{7}\hlc[pink!100]{5}\hlc[pink!100]{,}\hlc[pink!100]{1}\hlc[pink!100]{3}\hlc[pink!100]{3}\hlc[pink!100]{ }\hlc[pink!100]{ of}\hlc[pink!100]{ the}\hlc[pink!100]{ }\hlc[pink!100]{1}\hlc[pink!100]{2}\hlc[pink!100]{.}\hlc[pink!100]{5}\hlc[pink!100]{ million}\hlc[pink!100]{ people}\hlc[pink!100]{ living}\hlc[pink!100]{ in}\hlc[pink!100]{ New}\hlc[pink!100]{ York}\hlc[pink!100]{ State}\hlc[pink!100]{ live}\hlc[pink!100]{ in}\hlc[pink!100]{ New}\hlc[pink!100]{ York}\hlc[pink!100]{ City}\hlc[pink!100]{ } \newline
+ \hlc[pink!100]{ }\hlc[pink!100]{8}\hlc[pink!100]{.}\hlc[pink!100]{4}\hlc[pink!100]{ million}\hlc[pink!100]{ people}\hlc[pink!100]{ live}\hlc[pink!100]{ in}\hlc[pink!100]{ New}\hlc[pink!100]{ York}\hlc[pink!100]{ City}\hlc[pink!100]{.}\hlc[pink!100]{ } \newline
+ \hlc[pink!100]{ }\hlc[pink!100]{8}\hlc[pink!100]{,}\hlc[pink!100]{1}\hlc[pink!100]{7}\hlc[pink!100]{5}\hlc[pink!100]{,}\hlc[pink!100]{1}\hlc[pink!100]{3}\hlc[pink!100]{3}\hlc[pink!100]{ in}\hlc[pink!100]{ }\hlc[pink!100]{2}\hlc[pink!100]{0}\hlc[pink!100]{1}\hlc[pink!100]{0}\hlc[pink!100]{ and}\hlc[pink!100]{ }\hlc[pink!100]{8}\hlc[pink!100]{,}\hlc[pink!100]{4}\hlc[pink!100]{9}\hlc[pink!100]{1}\hlc[pink!100]{,}\hlc[pink!100]{0}\hlc[pink!100]{7}\hlc[pink!100]{9}\hlc[pink!100]{ in}\hlc[pink!100]{ }\hlc[pink!100]{2}\hlc[pink!100]{0}\hlc[pink!100]{1}\hlc[pink!100]{4}\hlc[pink!100]{.}\hlc[pink!100]{ [}\hlc[pink!100]{1}\hlc[pink!100]{7}\hlc[pink!100]{]}\hlc[pink!100]{ New}\hlc[pink!100]{ York}\hlc[pink!100]{ City}\hlc[pink!100]{ has}\hlc[pink!100]{ the}\hlc[pink!100]{ largest}\hlc[pink!100]{ city}\hlc[pink!100]{-}\hlc[pink!100]{wide}\hlc[pink!100]{ metropolitan}\hlc[pink!100]{ population}\hlc[pink!100]{ in}\hlc[pink!100]{ the}\hlc[pink!100]{ United}\hlc[pink!100]{ States}\hlc[pink!100]{,}\hlc[pink!100]{ at}\hlc[pink!100]{ }\hlc[pink!100]{2}\hlc[pink!100]{0}\hlc[pink!100]{.}\hlc[pink!100]{3}\hlc[pink!100]{ million}\hlc[pink!100]{ people}\hlc[pink!100]{.}\hlc[pink!100]{ [}\hlc[pink!100]{1}\hlc[pink!100]{8}\hlc[pink!100]{]}\hlc[pink!100]{ The}\hlc[pink!100]{ United}\hlc[pink!100]{ States}\hlc[pink!100]{ Census}\hlc[pink!100]{ Bureau}\hlc[pink!100]{ estimates}\hlc[pink!100]{ that}\hlc[pink!100]{ the}\hlc[pink!100]{ New}\hlc[pink!100]{ York}\hlc[pink!100]{ City}\hlc[pink!100]{ metropolitan}\hlc[pink!100]{ area}\hlc[pink!100]{ has}\hlc[pink!100]{ a}\hlc[pink!100]{ population}\hlc[pink!100]{ of}\hlc[pink!100]{ }\hlc[pink!100]{2}\hlc[pink!100]{0}\hlc[pink!100]{.}\hlc[pink!100]{9}\hlc[pink!100]{ million}\hlc[pink!100]{,}\hlc[pink!100]{ making}\hlc[pink!100]{ it}\hlc[pink!100]{ the}\hlc[pink!100]{ most}\hlc[pink!100]{ populous}\hlc[pink!100]{ metropolitan}\hlc[pink!100]{ area}\hlc[pink!100]{ in}\hlc[pink!100]{ the}\hlc[pink!100]{ United}\hlc[pink!100]{ States}\hlc[pink!100]{.}\hlc[pink!100]{ [}\hlc[pink!100]{1}\hlc[pink!100]{9}\hlc[pink!100]{]}\hlc[pink!100]{ The}\hlc[pink!100]{ New}\hlc[pink!100]{ York}\hlc[pink!100]{ City}\hlc[pink!100]{ metropolitan}\hlc[pink!100]{ area}\hlc[pink!100]{ is}\hlc[pink!100]{ the}\hlc[pink!100]{ fourth}\hlc[pink!100]{-}\hlc[pink!100]{largest}\hlc[pink!100]{ in}\hlc[pink!100]{ the}\hlc[pink!100]{ United}\hlc[pink!100]{ States}\hlc[pink!100]{ by}\hlc[pink!100]{ population}\hlc[pink!100]{,}\hlc[pink!100]{ behind}\hlc[pink!100]{ Los}\hlc[pink!100]{ Angeles} \newline
+ \hlc[pink!100]{ }\hlc[pink!100]{8}\hlc[pink!100]{.}\hlc[pink!100]{4}\hlc[pink!100]{9}\hlc[pink!100]{1}\hlc[pink!100]{ million}\hlc[pink!100]{ according}\hlc[pink!100]{ to}\hlc[pink!100]{ this}\hlc[pink!100]{ source}\hlc[pink!100]{.}\hlc[pink!100]{<0x0A>}\hlc[pink!100]{<0x0A>}\hlc[pink!100]{https}\hlc[pink!100]{://}\hlc[pink!100]{en}\hlc[pink!100]{.}\hlc[pink!100]{wikipedia}\hlc[pink!100]{.}\hlc[pink!100]{org}\hlc[pink!100]{/}\hlc[pink!100]{wiki}\hlc[pink!100]{/}\hlc[pink!100]{New}\hlc[pink!100]{\_}\hlc[pink!100]{York}\hlc[pink!100]{\_}\hlc[pink!100]{City} \\
\textbf{Guided top-p (=0.1)} \\
+ \hlc[pink!71]{ }\hlc[pink!49]{8}\hlc[pink!43]{.}\hlc[pink!0]{4}\hlc[pink!0]{9}\hlc[pink!0]{1}\hlc[pink!5]{ million}\hlc[pink!81]{ people}\hlc[pink!51]{ live}\hlc[pink!97]{ in}\hlc[pink!93]{ New}\hlc[pink!99]{ York}\hlc[pink!98]{ City}\hlc[pink!82]{ }  [x2] \newline
+ \hlc[pink!71]{ }\hlc[pink!51]{8}\hlc[pink!41]{.}\hlc[pink!0]{4}\hlc[pink!0]{9}\hlc[pink!0]{1}\hlc[pink!7]{ million}\hlc[pink!80]{ people}\hlc[pink!51]{ } [x4] \newline
+ \hlc[pink!71]{ }\hlc[pink!51]{8}\hlc[pink!41]{,}\hlc[pink!0]{4}\hlc[pink!0]{9}\hlc[pink!0]{1}\hlc[pink!94]{,}\hlc[pink!0]{0}\hlc[pink!0]{7}\hlc[pink!0]{9}\hlc[pink!40]{ people}\hlc[pink!79]{ live}\hlc[pink!91]{ in}\hlc[pink!93]{ New}\hlc[pink!99]{ York}\hlc[pink!98]{ City}\hlc[pink!25]{ } \newline
+ \hlc[pink!71]{ }\hlc[pink!51]{8}\hlc[pink!41]{.}\hlc[pink!0]{4}\hlc[pink!0]{9}\hlc[pink!0]{1}\hlc[pink!7]{ million}\hlc[pink!80]{ people}\hlc[pink!51]{ live}\hlc[pink!97]{ in}\hlc[pink!93]{ NYC}\hlc[pink!70]{.}\hlc[pink!3]{ } \newline
+ \hlc[pink!71]{ About}\hlc[pink!90]{ }\hlc[pink!39]{8}\hlc[pink!35]{.}\hlc[pink!13]{4}\hlc[pink!0]{ million}\hlc[pink!86]{ people}\hlc[pink!85]{ (}\hlc[pink!30]{8}\hlc[pink!88]{,}\hlc[pink!21]{4}\hlc[pink!0]{9}\hlc[pink!0]{1}\hlc[pink!92]{,}\hlc[pink!0]{0}\hlc[pink!0]{7}\hlc[pink!0]{9}\hlc[pink!30]{)}\hlc[pink!77]{ live}\hlc[pink!96]{ in}\hlc[pink!89]{ the}\hlc[pink!18]{ city}\hlc[pink!66]{,}\hlc[pink!29]{ making}\hlc[pink!97]{ it}\hlc[pink!97]{ one}\hlc[pink!99]{ of}\hlc[pink!98]{ the}\hlc[pink!88]{ most}\hlc[pink!96]{ densely}\hlc[pink!99]{ populated}\hlc[pink!62]{ cities}\hlc[pink!99]{ in}\hlc[pink!96]{ the}\hlc[pink!54]{ world}\hlc[pink!83]{ and}\hlc[pink!95]{ one}\hlc[pink!99]{ of}\hlc[pink!97]{ the}\hlc[pink!94]{ most}\hlc[pink!82]{ populous}\hlc[pink!68]{ in}\hlc[pink!86]{ the}\hlc[pink!92]{ United}\hlc[pink!99]{ States}\hlc[pink!86]{.}\hlc[pink!0]{ } \newline
+ \hlc[pink!71]{ How}\hlc[pink!87]{ to}\hlc[pink!67]{ find}\hlc[pink!91]{ out}\hlc[pink!88]{ }
\\
\hline
\end{tabular}
\caption{Prediction examples of the QReCC conversational QA task.}
\label{tab:example_conversational_qa_2}
\end{table*}

\begin{table*}
\centering
\begin{tabular}{p{0.96\linewidth}}
\hline
\textbf{Query} \\
Fact: Iglesias starred alongside Antonio Banderas , Salma Hayek and Johnny Depp in the Robert Rodriguez film Once Upon a Time in Mexico , in which he played the well-spoken gun-wielding Lorenzo. [57] In 2007, he had a guest appearance in the TV comedy Two and a Half Men as a carpenter/handyman. He also guest starred as Gael, an Argentinean guitar playing/surfer/massage therapist love interest of Robin in season 3 of the CBS hit TV show How I Met Your Mother . [58] Iglesias also played the part of an evil Roman emperor in a Pepsi advertising in 2004. [59] He's also starred in commercials for Tommy Hilfiger , Doritos and for Viceroy watches. Personal life [ edit ] Anna Kournikova In late 2001, Enrique Iglesias started a relationship with Russian tennis player Anna Kournikova . They reportedly split in October 2013 but have since reconciled. [60] The couple had twins, Nicholas and Lucy, on Saturday, 16 December 2017, at South Miami Hospital. [61] In 2003, Iglesias received surgery to remove a circular mole from the right side of his face, citing concerns that over time it could become cancerous. [62] Iglesias lives an active lifestyle and loves playing sports, surfing, and swimming in the ocean. In a 2014 interview, he stated: "I actually don’t have a specific routine. I mostly do push-ups and bench presses when I am at home. When I am on the road, I go to the gym to exercise. My shows are very active and provide a lot of cardio exercise." [63] 
\newline \newline
[eot] 0 -1 0 Where was Enrique Iglesias born? [eot] 1 0 1 Enrique Iglesias was born in Madrid, Spain. [eot] 2 1 0 Who were Enrique Iglesias' parents? [eot] 3 2 1 Spanish singer Julio Iglesias and Filipina socialite and magazine journalist Isabel Preysler [eot] 4 3 0 Was Enrique Inglesias married? [eot] 5 4 1 In late 2001, Enrique Iglesias started a relationship with Russian tennis player Anna Kournikova. [eot] 6 5 0 Did Enrique Iglesias have any children? [eot] 7 6 1
\\
\textbf{Greedy algorithm} \\
+ \hlc[pink!0]{ Yes}\hlc[pink!0]{,}\hlc[pink!0]{ he}\hlc[pink!0]{ has}\hlc[pink!0]{ two}\hlc[pink!0]{ children}\hlc[pink!0]{.}\hlc[pink!0]{ } [x8] \newline
+ \hlc[pink!0]{ Yes}\hlc[pink!0]{,}\hlc[pink!0]{ he}\hlc[pink!0]{ has}\hlc[pink!0]{ two}\hlc[pink!0]{ children}\hlc[pink!0]{ with}\hlc[pink!0]{ Anna}\hlc[pink!0]{ Kour}\hlc[pink!0]{nikova}\hlc[pink!0]{.}\hlc[pink!0]{ } [x2]
\\
\textbf{Baseline top-p (=0.3)} \\
+ \hlc[pink!100]{ Yes}\hlc[pink!100]{,}\hlc[pink!100]{ he}\hlc[pink!100]{ has}\hlc[pink!100]{ two}\hlc[pink!100]{ children}\hlc[pink!100]{.}\hlc[pink!100]{ } [x3] \newline
+ \hlc[pink!100]{ Yes}\hlc[pink!100]{,}\hlc[pink!100]{ he}\hlc[pink!100]{ has}\hlc[pink!100]{ two}\hlc[pink!100]{ children}\hlc[pink!100]{ with}\hlc[pink!100]{ Anna}\hlc[pink!100]{ Kour}\hlc[pink!100]{nikova}\hlc[pink!100]{.}\hlc[pink!100]{ } [x5] \newline
+ \hlc[pink!100]{ He}\hlc[pink!100]{ has}\hlc[pink!100]{ two}\hlc[pink!100]{ children}\hlc[pink!100]{,}\hlc[pink!100]{ Nicholas}\hlc[pink!100]{ and}\hlc[pink!100]{ Lucy}\hlc[pink!100]{.}\hlc[pink!100]{ } \newline
+ \hlc[pink!100]{ He}\hlc[pink!100]{ has}\hlc[pink!100]{ two}\hlc[pink!100]{ children}\hlc[pink!100]{ with}\hlc[pink!100]{ Anna}\hlc[pink!100]{ Kour}\hlc[pink!100]{nikova}\hlc[pink!100]{,}\hlc[pink!100]{ Nicholas}\hlc[pink!100]{ and}\hlc[pink!100]{ Lucy}\hlc[pink!100]{.}\hlc[pink!100]{ }
\\
\textbf{Baseline top-p (=0.7)} \\
+ \hlc[pink!100]{ Yes}\hlc[pink!100]{,}\hlc[pink!100]{ he}\hlc[pink!100]{ has}\hlc[pink!100]{ two}\hlc[pink!100]{ children}\hlc[pink!100]{ with}\hlc[pink!100]{ Anna}\hlc[pink!100]{ Kour}\hlc[pink!100]{nikova}\hlc[pink!100]{.}\hlc[pink!100]{ } [x2] \newline
+ \hlc[pink!100]{ He}\hlc[pink!100]{ has}\hlc[pink!100]{ two}\hlc[pink!100]{ children}\hlc[pink!100]{.}\hlc[pink!100]{ } \newline
+ \hlc[pink!100]{ He}\hlc[pink!100]{ has}\hlc[pink!100]{ a}\hlc[pink!100]{ daughter}\hlc[pink!100]{,}\hlc[pink!100]{ Natalia}\hlc[pink!100]{ and}\hlc[pink!100]{ a}\hlc[pink!100]{ son}\hlc[pink!100]{,}\hlc[pink!100]{ Nicholas}\hlc[pink!100]{.}\hlc[pink!100]{ } \newline
+ \hlc[pink!100]{ He}\hlc[pink!100]{ has}\hlc[pink!100]{ a}\hlc[pink!100]{ daughter}\hlc[pink!100]{ named}\hlc[pink!100]{ Natalia}\hlc[pink!100]{.}\hlc[pink!100]{ } \newline
+ \hlc[pink!100]{ He}\hlc[pink!100]{ had}\hlc[pink!100]{ twins}\hlc[pink!100]{ with}\hlc[pink!100]{ his}\hlc[pink!100]{ wife}\hlc[pink!100]{ Anna}\hlc[pink!100]{,}\hlc[pink!100]{ Nicholas}\hlc[pink!100]{ and}\hlc[pink!100]{ Lucy}\hlc[pink!100]{.}\hlc[pink!100]{ } \newline
+ \hlc[pink!100]{ Two}\hlc[pink!100]{ sons}\hlc[pink!100]{:}\hlc[pink!100]{ Nicholas}\hlc[pink!100]{ (}\hlc[pink!100]{born}\hlc[pink!100]{ }\hlc[pink!100]{2}\hlc[pink!100]{0}\hlc[pink!100]{0}\hlc[pink!100]{3}\hlc[pink!100]{)}\hlc[pink!100]{ and}\hlc[pink!100]{ Lucy}\hlc[pink!100]{ (}\hlc[pink!100]{born}\hlc[pink!100]{ }\hlc[pink!100]{2}\hlc[pink!100]{0}\hlc[pink!100]{0}\hlc[pink!100]{6}\hlc[pink!100]{)}\hlc[pink!100]{ with}\hlc[pink!100]{ his}\hlc[pink!100]{ Russian}\hlc[pink!100]{ tennis}\hlc[pink!100]{ player}\hlc[pink!100]{ girlfriend}\hlc[pink!100]{ Anna}\hlc[pink!100]{ Kour}\hlc[pink!100]{nikova}\hlc[pink!100]{.}\hlc[pink!100]{ } \newline
+ \hlc[pink!100]{ Yes}\hlc[pink!100]{,}\hlc[pink!100]{ two}\hlc[pink!100]{.}\hlc[pink!100]{ Nicholas}\hlc[pink!100]{ and}\hlc[pink!100]{ Lucy}\hlc[pink!100]{.}\hlc[pink!100]{ } \newline
+ \hlc[pink!100]{ Two}\hlc[pink!100]{ daughters}\hlc[pink!100]{,}\hlc[pink!100]{ Maria}\hlc[pink!100]{ Isabel}\hlc[pink!100]{ and}\hlc[pink!100]{ Sofia}\hlc[pink!100]{.}\hlc[pink!100]{ } \newline
+ \hlc[pink!100]{ Two}\hlc[pink!100]{ children}\hlc[pink!100]{:}\hlc[pink!100]{ Nicholas}\hlc[pink!100]{ and}\hlc[pink!100]{ Lucy}\hlc[pink!100]{ }
\\
\textbf{Guided top-p (=0.1)} \\
+ \hlc[pink!3]{ Yes}\hlc[pink!87]{.}\hlc[pink!4]{ } [x2] \newline
+ \hlc[pink!3]{ Yes}\hlc[pink!87]{,}\hlc[pink!11]{ he}\hlc[pink!95]{ has}\hlc[pink!12]{ two}\hlc[pink!38]{ children}\hlc[pink!10]{ with}\hlc[pink!23]{ Anna}\hlc[pink!77]{ Kour}\hlc[pink!99]{nikova}\hlc[pink!49]{.}\hlc[pink!14]{ } [x2] \newline
+ \hlc[pink!3]{ Yes}\hlc[pink!87]{,}\hlc[pink!11]{ he}\hlc[pink!95]{ had}\hlc[pink!0]{ two}\hlc[pink!34]{ children}\hlc[pink!29]{ with}\hlc[pink!27]{ Anna}\hlc[pink!73]{ Kour}\hlc[pink!99]{nikova}\hlc[pink!56]{.}\hlc[pink!17]{ } [x3] \newline
+ \hlc[pink!3]{ Yes}\hlc[pink!87]{,}\hlc[pink!11]{ he}\hlc[pink!95]{ did}\hlc[pink!90]{,}\hlc[pink!3]{ he}\hlc[pink!96]{ had}\hlc[pink!0]{ two}\hlc[pink!18]{ children}\hlc[pink!15]{ with}\hlc[pink!41]{ Anna}\hlc[pink!76]{ Kour}\hlc[pink!99]{nikova}\hlc[pink!50]{.}\hlc[pink!27]{ } \newline
+ \hlc[pink!3]{ Yes}\hlc[pink!87]{,}\hlc[pink!11]{ he}\hlc[pink!95]{ has}\hlc[pink!12]{ two}\hlc[pink!38]{ children}\hlc[pink!10]{,}\hlc[pink!0]{ Nicholas}\hlc[pink!5]{ and}\hlc[pink!0]{ Lucy}\hlc[pink!77]{.}\hlc[pink!81]{ } [x2]
\\
\hline
\end{tabular}
\caption{Prediction examples of the QReCC conversational QA task.}
\label{tab:example_conversational_qa_3}
\end{table*}

%% file: example_summarization.tex
\begin{table*}
\centering
\begin{tabular}{p{0.96\linewidth}}
\hline
\textbf{Query} \\
Farmer Robert Fidler 66, built the four-bedroom home near Redhill, Surrey, without planning permission. He was told on Monday he would be jailed for his "defiance" if the property was not demolished by June. Reigate and Banstead Borough Council said it would send in the bulldozers if Mr Fidler did not knock down the house. Mr Fidler would have to repay the council's costs. "We don't have choice - ultimately we have to ensure that these notices are complied with," said planning committee chairman Mark Brunt. "On 6 June we expect the site to be clear and the building to be gone." Mr Fidler told the High Court he had sold the house at Honeycrock Farm in Salfords to an Indian businessman and that the injunction ordering demolition was invalid. The house was built on greenbelt land between 1999 and 2006. The council first ordered its demolition in 2007. Mr Fidler claimed the council wanted to destroy his life and that its case was based on lies and deception. The council said there was another house on the farm where Mr Fidler and his family used to live. "I urge him to come and continue to talk to the council and look at realistic options for providing accommodation for his family on the site," said Mr Brunt. Planning law expert Martin Goodall said Mr Fidler had reached the end of the road and would have to demolish the house. "There are very strong policies against building isolated houses in the greenbelt or open countryside and local authorities cannot allow it," he said. \\
\textbf{Greedy algorithm} \\
+ A council has ordered demolition of a house in the greenbelt of Surrey. [x5] \\
\textbf{Baseline top-p (=0.4)} \\
+ \hlc[pink!100]{ A}\hlc[pink!100]{ }\hlc[pink!100]{council}\hlc[pink!100]{ has}\hlc[pink!100]{ }\hlc[pink!100]{ordered}\hlc[pink!100]{ }\hlc[pink!100]{demoli}\hlc[pink!100]{tion}\hlc[pink!100]{ of}\hlc[pink!100]{ }\hlc[pink!100]{a}\hlc[pink!100]{ house}\hlc[pink!100]{ on}\hlc[pink!100]{ green}\hlc[pink!100]{belt}\hlc[pink!100]{ land}\hlc[pink!100]{ in}\hlc[pink!100]{ }\hlc[pink!100]{Hampshire}\hlc[pink!100]{.}\hlc[pink!100]{}

+ \hlc[pink!100]{ A}\hlc[pink!100]{ farmer}\hlc[pink!100]{ has}\hlc[pink!100]{ }\hlc[pink!100]{told}\hlc[pink!100]{ the}\hlc[pink!100]{ High}\hlc[pink!100]{ Court}\hlc[pink!100]{ he}\hlc[pink!100]{ }\hlc[pink!100]{would}\hlc[pink!100]{ be}\hlc[pink!100]{ ja}\hlc[pink!100]{iled}\hlc[pink!100]{ if}\hlc[pink!100]{ he}\hlc[pink!100]{ did}\hlc[pink!100]{ not}\hlc[pink!100]{ dem}\hlc[pink!100]{olish}\hlc[pink!100]{ his}\hlc[pink!100]{ house}\hlc[pink!100]{.}\hlc[pink!100]{}

+ \hlc[pink!100]{ A}\hlc[pink!100]{ }\hlc[pink!100]{council}\hlc[pink!100]{ has}\hlc[pink!100]{ }\hlc[pink!100]{told}\hlc[pink!100]{ }\hlc[pink!100]{a}\hlc[pink!100]{ farmer}\hlc[pink!100]{ to}\hlc[pink!100]{ dem}\hlc[pink!100]{olish}\hlc[pink!100]{ his}\hlc[pink!100]{ house}\hlc[pink!100]{ on}\hlc[pink!100]{ green}\hlc[pink!100]{belt}\hlc[pink!100]{ land}\hlc[pink!100]{.}\hlc[pink!100]{}

+ \hlc[pink!100]{ A}\hlc[pink!100]{ farmer}\hlc[pink!100]{ has}\hlc[pink!100]{ }\hlc[pink!100]{told}\hlc[pink!100]{ the}\hlc[pink!100]{ High}\hlc[pink!100]{ Court}\hlc[pink!100]{ he}\hlc[pink!100]{ }\hlc[pink!100]{would}\hlc[pink!100]{ be}\hlc[pink!100]{ ja}\hlc[pink!100]{iled}\hlc[pink!100]{ if}\hlc[pink!100]{ he}\hlc[pink!100]{ did}\hlc[pink!100]{ not}\hlc[pink!100]{ dem}\hlc[pink!100]{olish}\hlc[pink!100]{ his}\hlc[pink!100]{ house}\hlc[pink!100]{ on}\hlc[pink!100]{ green}\hlc[pink!100]{belt}\hlc[pink!100]{ land}\hlc[pink!100]{.}\hlc[pink!100]{}

+ \hlc[pink!100]{ A}\hlc[pink!100]{ farmer}\hlc[pink!100]{ has}\hlc[pink!100]{ }\hlc[pink!100]{taken}\hlc[pink!100]{ legal}\hlc[pink!100]{ action}\hlc[pink!100]{ }\hlc[pink!100]{against}\hlc[pink!100]{ }\hlc[pink!100]{a}\hlc[pink!100]{ }\hlc[pink!100]{council}\hlc[pink!100]{ }\hlc[pink!100]{which}\hlc[pink!100]{ has}\hlc[pink!100]{ }\hlc[pink!100]{ordered}\hlc[pink!100]{ his}\hlc[pink!100]{ house}\hlc[pink!100]{ to}\hlc[pink!100]{ be}\hlc[pink!100]{ dem}\hlc[pink!100]{olished}\hlc[pink!100]{.}\hlc[pink!100]{}
\\
\textbf{Baseline top-p (=0.7)} \\
+ \hlc[pink!100]{ A}\hlc[pink!100]{ farmer}\hlc[pink!100]{ in}\hlc[pink!100]{ }\hlc[pink!100]{Hampshire}\hlc[pink!100]{ has}\hlc[pink!100]{ been}\hlc[pink!100]{ }\hlc[pink!100]{told}\hlc[pink!100]{ he}\hlc[pink!100]{ }\hlc[pink!100]{would}\hlc[pink!100]{ be}\hlc[pink!100]{ ja}\hlc[pink!100]{iled}\hlc[pink!100]{ for}\hlc[pink!100]{ his}\hlc[pink!100]{ "}\hlc[pink!100]{de}\hlc[pink!100]{fianc}\hlc[pink!100]{e}\hlc[pink!100]{"}\hlc[pink!100]{ if}\hlc[pink!100]{ he}\hlc[pink!100]{ did}\hlc[pink!100]{ not}\hlc[pink!100]{ dem}\hlc[pink!100]{olish}\hlc[pink!100]{ his}\hlc[pink!100]{ house}\hlc[pink!100]{.}\hlc[pink!100]{}

+ \hlc[pink!100]{ An}\hlc[pink!100]{ }\hlc[pink!100]{Isle}\hlc[pink!100]{ of}\hlc[pink!100]{ Man}\hlc[pink!100]{ farmer}\hlc[pink!100]{ has}\hlc[pink!100]{ been}\hlc[pink!100]{ }\hlc[pink!100]{told}\hlc[pink!100]{ he}\hlc[pink!100]{ }\hlc[pink!100]{would}\hlc[pink!100]{ be}\hlc[pink!100]{ ja}\hlc[pink!100]{iled}\hlc[pink!100]{ if}\hlc[pink!100]{ his}\hlc[pink!100]{ green}\hlc[pink!100]{belt}\hlc[pink!100]{ home}\hlc[pink!100]{ was}\hlc[pink!100]{ not}\hlc[pink!100]{ dem}\hlc[pink!100]{olished}\hlc[pink!100]{ by}\hlc[pink!100]{ }\hlc[pink!100]{a}\hlc[pink!100]{ }\hlc[pink!100]{council}\hlc[pink!100]{.}\hlc[pink!100]{}

+ \hlc[pink!100]{ A}\hlc[pink!100]{ }\hlc[pink!100]{council}\hlc[pink!100]{ has}\hlc[pink!100]{ }\hlc[pink!100]{given}\hlc[pink!100]{ the}\hlc[pink!100]{ land}\hlc[pink!100]{lord}\hlc[pink!100]{ }\hlc[pink!100]{notice}\hlc[pink!100]{ to}\hlc[pink!100]{ dem}\hlc[pink!100]{olish}\hlc[pink!100]{ his}\hlc[pink!100]{ house}\hlc[pink!100]{ after}\hlc[pink!100]{ it}\hlc[pink!100]{ }\hlc[pink!100]{ordered}\hlc[pink!100]{ it}\hlc[pink!100]{ to}\hlc[pink!100]{ be}\hlc[pink!100]{ dem}\hlc[pink!100]{olished}\hlc[pink!100]{.}\hlc[pink!100]{}

+ \hlc[pink!100]{ }\hlc[pink!100]{Resident}\hlc[pink!100]{s}\hlc[pink!100]{ of}\hlc[pink!100]{ }\hlc[pink!100]{a}\hlc[pink!100]{ farm}\hlc[pink!100]{ in}\hlc[pink!100]{ south}\hlc[pink!100]{ London}\hlc[pink!100]{ have}\hlc[pink!100]{ been}\hlc[pink!100]{ }\hlc[pink!100]{told}\hlc[pink!100]{ the}\hlc[pink!100]{y}\hlc[pink!100]{ have}\hlc[pink!100]{ to}\hlc[pink!100]{ dem}\hlc[pink!100]{olish}\hlc[pink!100]{ }\hlc[pink!100]{a}\hlc[pink!100]{ house}\hlc[pink!100]{ that}\hlc[pink!100]{ the}\hlc[pink!100]{y}\hlc[pink!100]{ }\hlc[pink!100]{believe}\hlc[pink!100]{d}\hlc[pink!100]{ was}\hlc[pink!100]{ }\hlc[pink!100]{built}\hlc[pink!100]{ }\hlc[pink!100]{without}\hlc[pink!100]{ }\hlc[pink!100]{planning}\hlc[pink!100]{ }\hlc[pink!100]{permission}\hlc[pink!100]{.}\hlc[pink!100]{}

+ \hlc[pink!100]{ A}\hlc[pink!100]{ man}\hlc[pink!100]{ has}\hlc[pink!100]{ be}\hlc[pink!100]{gun}\hlc[pink!100]{ }\hlc[pink!100]{a}\hlc[pink!100]{ legal}\hlc[pink!100]{ battle}\hlc[pink!100]{ }\hlc[pink!100]{against}\hlc[pink!100]{ }\hlc[pink!100]{a}\hlc[pink!100]{ }\hlc[pink!100]{council}\hlc[pink!100]{ }\hlc[pink!100]{because}\hlc[pink!100]{ he}\hlc[pink!100]{ had}\hlc[pink!100]{ been}\hlc[pink!100]{ }\hlc[pink!100]{told}\hlc[pink!100]{ to}\hlc[pink!100]{ dem}\hlc[pink!100]{olish}\hlc[pink!100]{ his}\hlc[pink!100]{ house}\hlc[pink!100]{ on}\hlc[pink!100]{ green}\hlc[pink!100]{belt}\hlc[pink!100]{ land}\hlc[pink!100]{.}\hlc[pink!100]{} \\
\textbf{Guided top-p (=1.0)} \\
+ \hlc[pink!15]{ A}\hlc[pink!14]{ }\hlc[pink!3]{council}\hlc[pink!28]{ has}\hlc[pink!69]{ order}\hlc[pink!87]{ed}\hlc[pink!51]{ }\hlc[pink!8]{demoli}\hlc[pink!99]{tion}\hlc[pink!80]{ of}\hlc[pink!34]{ }\hlc[pink!88]{a}\hlc[pink!32]{ house}\hlc[pink!55]{ in}\hlc[pink!23]{ the}\hlc[pink!0]{ green}\hlc[pink!4]{belt}\hlc[pink!61]{ of}\hlc[pink!27]{ Sur}\hlc[pink!99]{rey}\hlc[pink!84]{.}\hlc[pink!90]{}

+ \hlc[pink!15]{ A}\hlc[pink!14]{ }\hlc[pink!3]{council}\hlc[pink!28]{ has}\hlc[pink!69]{ been}\hlc[pink!60]{ }\hlc[pink!19]{told}\hlc[pink!51]{ to}\hlc[pink!15]{ send}\hlc[pink!8]{ bull}\hlc[pink!97]{do}\hlc[pink!96]{zers}\hlc[pink!69]{ to}\hlc[pink!17]{ dem}\hlc[pink!0]{olish}\hlc[pink!60]{ }\hlc[pink!59]{a}\hlc[pink!24]{ house}\hlc[pink!52]{ on}\hlc[pink!4]{ green}\hlc[pink!2]{belt}\hlc[pink!0]{ land}\hlc[pink!63]{.}\hlc[pink!81]{}

+ \hlc[pink!15]{ A}\hlc[pink!14]{ }\hlc[pink!3]{council}\hlc[pink!28]{ has}\hlc[pink!69]{ warn}\hlc[pink!99]{ed}\hlc[pink!58]{ }\hlc[pink!39]{a}\hlc[pink!6]{ farmer}\hlc[pink!75]{ to}\hlc[pink!26]{ dem}\hlc[pink!9]{olish}\hlc[pink!43]{ his}\hlc[pink!32]{ house}\hlc[pink!71]{ on}\hlc[pink!2]{ green}\hlc[pink!1]{belt}\hlc[pink!0]{ land}\hlc[pink!64]{.}\hlc[pink!89]{}

+ \hlc[pink!15]{ A}\hlc[pink!14]{ }\hlc[pink!3]{council}\hlc[pink!28]{ has}\hlc[pink!69]{ }\hlc[pink!34]{ordered}\hlc[pink!72]{ }\hlc[pink!9]{demoli}\hlc[pink!99]{tion}\hlc[pink!84]{ of}\hlc[pink!40]{ }\hlc[pink!91]{a}\hlc[pink!34]{ house}\hlc[pink!54]{ that}\hlc[pink!74]{ is}\hlc[pink!80]{ on}\hlc[pink!1]{ green}\hlc[pink!0]{belt}\hlc[pink!0]{ land}\hlc[pink!70]{.}\hlc[pink!83]{}

+ \hlc[pink!15]{ A}\hlc[pink!14]{ }\hlc[pink!3]{council}\hlc[pink!28]{ has}\hlc[pink!69]{ }\hlc[pink!34]{told}\hlc[pink!57]{ }\hlc[pink!56]{a}\hlc[pink!2]{ farmer}\hlc[pink!61]{ that}\hlc[pink!81]{ he}\hlc[pink!74]{ }\hlc[pink!73]{would}\hlc[pink!55]{ be}\hlc[pink!1]{ ja}\hlc[pink!99]{iled}\hlc[pink!73]{ if}\hlc[pink!81]{ he}\hlc[pink!56]{ did}\hlc[pink!77]{ not}\hlc[pink!3]{ dem}\hlc[pink!82]{olish}\hlc[pink!83]{ his}\hlc[pink!41]{ house}\hlc[pink!75]{ in}\hlc[pink!24]{ }\hlc[pink!59]{a}\hlc[pink!0]{ green}\hlc[pink!0]{belt}\hlc[pink!67]{.}\hlc[pink!87]{}
\\
\hline
\end{tabular}
\caption{Prediction examples of the XLSum summarization task.}
\label{tab:example_summarization_1}
\end{table*}

\begin{table*}
\centering
\begin{tabular}{p{0.96\linewidth}}
\hline
\textbf{Query} \\
A total of 84 people died in the last 24 hours, Governor Andrew Cuomo said on Saturday, compared with 109 a day before. During the height of the outbreak in April, more than 1,000 people a day were losing their lives in worst-hit US state. "In my head, I was always looking to get under 100," Mr Cuomo said. "It doesn't do good for any of those 84 families that are feeling the pain," he said at his daily briefing, but added that the drop was a sign of "real progress". Mr Cuomo announced on Friday that groups of up to 10 people could gather "for any lawful purpose" anywhere in the state, including New York City. But, he added: "If you don't have to be with a group of 10 people don't be with a group of 10 people." New York state was once the epicentre of the US coronavirus outbreak, with more than 28,000 deaths, according to Johns Hopkins University. The US has the biggest death toll from Covid-19 at 96,000. The UK is second with more than 36,000. \\
\textbf{Greedy algorithm} \\
+ New York state has fallen below 100 deaths a day in the coronavirus outbreak, according to the governor. [x5] \\
\textbf{Baseline top-p (=0.4)} \\
+ \hlc[pink!100]{ New}\hlc[pink!100]{ York}\hlc[pink!100]{ state}\hlc[pink!100]{ has}\hlc[pink!100]{ }\hlc[pink!100]{fallen}\hlc[pink!100]{ short}\hlc[pink!100]{ of}\hlc[pink!100]{ the}\hlc[pink!100]{ 1,000}\hlc[pink!100]{ people}\hlc[pink!100]{ }\hlc[pink!100]{a}\hlc[pink!100]{ day}\hlc[pink!100]{ }\hlc[pink!100]{which}\hlc[pink!100]{ the}\hlc[pink!100]{ state}\hlc[pink!100]{ was}\hlc[pink!100]{ }\hlc[pink!100]{once}\hlc[pink!100]{ the}\hlc[pink!100]{ epi}\hlc[pink!100]{centre}\hlc[pink!100]{ of}\hlc[pink!100]{ the}\hlc[pink!100]{ coronavirus}\hlc[pink!100]{ out}\hlc[pink!100]{break}\hlc[pink!100]{.}\hlc[pink!100]{}

+ \hlc[pink!100]{ New}\hlc[pink!100]{ York}\hlc[pink!100]{ state}\hlc[pink!100]{ has}\hlc[pink!100]{ }\hlc[pink!100]{fallen}\hlc[pink!100]{ short}\hlc[pink!100]{ of}\hlc[pink!100]{ the}\hlc[pink!100]{ number}\hlc[pink!100]{ of}\hlc[pink!100]{ people}\hlc[pink!100]{ }\hlc[pink!100]{killed}\hlc[pink!100]{ by}\hlc[pink!100]{ coronavirus}\hlc[pink!100]{ in}\hlc[pink!100]{ the}\hlc[pink!100]{ last}\hlc[pink!100]{ 24}\hlc[pink!100]{ hours}\hlc[pink!100]{,}\hlc[pink!100]{ }\hlc[pink!100]{a}\hlc[pink!100]{ }\hlc[pink!100]{few}\hlc[pink!100]{ days}\hlc[pink!100]{ after}\hlc[pink!100]{ the}\hlc[pink!100]{ }\hlc[pink!100]{highest}\hlc[pink!100]{ number}\hlc[pink!100]{ of}\hlc[pink!100]{ death}\hlc[pink!100]{s}\hlc[pink!100]{.}\hlc[pink!100]{}

+ \hlc[pink!100]{ New}\hlc[pink!100]{ York}\hlc[pink!100]{ state}\hlc[pink!100]{ has}\hlc[pink!100]{ }\hlc[pink!100]{fallen}\hlc[pink!100]{ below}\hlc[pink!100]{ 100}\hlc[pink!100]{ people}\hlc[pink!100]{ }\hlc[pink!100]{a}\hlc[pink!100]{ day}\hlc[pink!100]{ in}\hlc[pink!100]{ the}\hlc[pink!100]{ last}\hlc[pink!100]{ 24}\hlc[pink!100]{ hours}\hlc[pink!100]{ }\hlc[pink!100]{during}\hlc[pink!100]{ the}\hlc[pink!100]{ coronavirus}\hlc[pink!100]{ pandemi}\hlc[pink!100]{c}\hlc[pink!100]{,}\hlc[pink!100]{ the}\hlc[pink!100]{ govern}\hlc[pink!100]{or}\hlc[pink!100]{ said}\hlc[pink!100]{.}\hlc[pink!100]{}

+ \hlc[pink!100]{ A}\hlc[pink!100]{ }\hlc[pink!100]{significant}\hlc[pink!100]{ drop}\hlc[pink!100]{ in}\hlc[pink!100]{ death}\hlc[pink!100]{s}\hlc[pink!100]{ in}\hlc[pink!100]{ New}\hlc[pink!100]{ York}\hlc[pink!100]{ state}\hlc[pink!100]{ has}\hlc[pink!100]{ been}\hlc[pink!100]{ }\hlc[pink!100]{reported}\hlc[pink!100]{,}\hlc[pink!100]{ }\hlc[pink!100]{a}\hlc[pink!100]{ }\hlc[pink!100]{chief}\hlc[pink!100]{ }\hlc[pink!100]{executive}\hlc[pink!100]{ said}\hlc[pink!100]{.}\hlc[pink!100]{}

+ \hlc[pink!100]{ New}\hlc[pink!100]{ York}\hlc[pink!100]{ state}\hlc[pink!100]{ has}\hlc[pink!100]{ been}\hlc[pink!100]{ slow}\hlc[pink!100]{ed}\hlc[pink!100]{ down}\hlc[pink!100]{ in}\hlc[pink!100]{ the}\hlc[pink!100]{ number}\hlc[pink!100]{ of}\hlc[pink!100]{ death}\hlc[pink!100]{s}\hlc[pink!100]{ from}\hlc[pink!100]{ coronavirus}\hlc[pink!100]{ }\hlc[pink!100]{during}\hlc[pink!100]{ the}\hlc[pink!100]{ last}\hlc[pink!100]{ 24}\hlc[pink!100]{ hours}\hlc[pink!100]{,}\hlc[pink!100]{ }\hlc[pink!100]{according}\hlc[pink!100]{ to}\hlc[pink!100]{ the}\hlc[pink!100]{ govern}\hlc[pink!100]{or}\hlc[pink!100]{.}\hlc[pink!100]{}
\\
\textbf{Baseline top-p (=0.7)} \\
+ \hlc[pink!100]{ New}\hlc[pink!100]{ York}\hlc[pink!100]{ State}\hlc[pink!100]{ has}\hlc[pink!100]{ had}\hlc[pink!100]{ }\hlc[pink!100]{a}\hlc[pink!100]{ }\hlc[pink!100]{significant}\hlc[pink!100]{ drop}\hlc[pink!100]{ in}\hlc[pink!100]{ death}\hlc[pink!100]{s}\hlc[pink!100]{ from}\hlc[pink!100]{ coronavirus}\hlc[pink!100]{ in}\hlc[pink!100]{ the}\hlc[pink!100]{ last}\hlc[pink!100]{ 24}\hlc[pink!100]{ hours}\hlc[pink!100]{,}\hlc[pink!100]{ }\hlc[pink!100]{according}\hlc[pink!100]{ to}\hlc[pink!100]{ the}\hlc[pink!100]{ state}\hlc[pink!100]{ govern}\hlc[pink!100]{or}\hlc[pink!100]{.}\hlc[pink!100]{}

+ \hlc[pink!100]{ New}\hlc[pink!100]{ York}\hlc[pink!100]{ state}\hlc[pink!100]{ has}\hlc[pink!100]{ had}\hlc[pink!100]{ }\hlc[pink!100]{a}\hlc[pink!100]{ drop}\hlc[pink!100]{ in}\hlc[pink!100]{ death}\hlc[pink!100]{s}\hlc[pink!100]{ from}\hlc[pink!100]{ coronavirus}\hlc[pink!100]{,}\hlc[pink!100]{ with}\hlc[pink!100]{ the}\hlc[pink!100]{ state}\hlc[pink!100]{ }\hlc[pink!100]{leadership}\hlc[pink!100]{ say}\hlc[pink!100]{ing}\hlc[pink!100]{ it}\hlc[pink!100]{ has}\hlc[pink!100]{ }\hlc[pink!100]{a}\hlc[pink!100]{ "}\hlc[pink!100]{real}\hlc[pink!100]{ progress}\hlc[pink!100]{"}\hlc[pink!100]{ to}\hlc[pink!100]{ be}\hlc[pink!100]{ under}\hlc[pink!100]{ 100}\hlc[pink!100]{.}\hlc[pink!100]{}

+ \hlc[pink!100]{ New}\hlc[pink!100]{ York}\hlc[pink!100]{ state}\hlc[pink!100]{ has}\hlc[pink!100]{ }\hlc[pink!100]{fallen}\hlc[pink!100]{ to}\hlc[pink!100]{ the}\hlc[pink!100]{ }\hlc[pink!100]{lowest}\hlc[pink!100]{ number}\hlc[pink!100]{ of}\hlc[pink!100]{ death}\hlc[pink!100]{s}\hlc[pink!100]{ from}\hlc[pink!100]{ coronavirus}\hlc[pink!100]{ }\hlc[pink!100]{-}\hlc[pink!100]{ in}\hlc[pink!100]{ the}\hlc[pink!100]{ last}\hlc[pink!100]{ 24}\hlc[pink!100]{ hours}\hlc[pink!100]{,}\hlc[pink!100]{ }\hlc[pink!100]{according}\hlc[pink!100]{ to}\hlc[pink!100]{ the}\hlc[pink!100]{ govern}\hlc[pink!100]{or}\hlc[pink!100]{.}\hlc[pink!100]{}

+ \hlc[pink!100]{ New}\hlc[pink!100]{ York}\hlc[pink!100]{ State}\hlc[pink!100]{'}\hlc[pink!100]{s}\hlc[pink!100]{ death}\hlc[pink!100]{ toll}\hlc[pink!100]{ from}\hlc[pink!100]{ Covid}\hlc[pink!100]{-19}\hlc[pink!100]{ has}\hlc[pink!100]{ }\hlc[pink!100]{fallen}\hlc[pink!100]{ below}\hlc[pink!100]{ 100}\hlc[pink!100]{,}\hlc[pink!100]{ the}\hlc[pink!100]{ govern}\hlc[pink!100]{or}\hlc[pink!100]{ said}\hlc[pink!100]{.}\hlc[pink!100]{}

+ \hlc[pink!100]{ A}\hlc[pink!100]{ de}\hlc[pink!100]{crease}\hlc[pink!100]{ in}\hlc[pink!100]{ death}\hlc[pink!100]{s}\hlc[pink!100]{ from}\hlc[pink!100]{ Covid}\hlc[pink!100]{-19}\hlc[pink!100]{ }\hlc[pink!100]{cases}\hlc[pink!100]{ in}\hlc[pink!100]{ New}\hlc[pink!100]{ York}\hlc[pink!100]{ State}\hlc[pink!100]{ is}\hlc[pink!100]{ }\hlc[pink!100]{a}\hlc[pink!100]{ sign}\hlc[pink!100]{ of}\hlc[pink!100]{ "}\hlc[pink!100]{real}\hlc[pink!100]{ progress}\hlc[pink!100]{"}\hlc[pink!100]{ by}\hlc[pink!100]{ }\hlc[pink!100]{a}\hlc[pink!100]{ govern}\hlc[pink!100]{or}\hlc[pink!100]{,}\hlc[pink!100]{ }\hlc[pink!100]{according}\hlc[pink!100]{ to}\hlc[pink!100]{ the}\hlc[pink!100]{ }\hlc[pink!100]{agency}\hlc[pink!100]{ }\hlc[pink!100]{CDC}\hlc[pink!100]{.}\hlc[pink!100]{}
\\
\textbf{Guided top-p (=1.0)} \\
+ \hlc[pink!6]{ New}\hlc[pink!48]{ York}\hlc[pink!3]{ state}\hlc[pink!23]{ has}\hlc[pink!33]{ }\hlc[pink!3]{fallen}\hlc[pink!26]{ below}\hlc[pink!7]{ 100}\hlc[pink!15]{ death}\hlc[pink!12]{s}\hlc[pink!56]{ from}\hlc[pink!33]{ coronavirus}\hlc[pink!62]{ }\hlc[pink!48]{during}\hlc[pink!60]{ the}\hlc[pink!56]{ last}\hlc[pink!13]{ 24}\hlc[pink!95]{ hours}\hlc[pink!67]{,}\hlc[pink!77]{ }\hlc[pink!83]{a}\hlc[pink!27]{ }\hlc[pink!64]{significant}\hlc[pink!11]{ drop}\hlc[pink!72]{ from}\hlc[pink!66]{ the}\hlc[pink!54]{ previous}\hlc[pink!14]{ day}\hlc[pink!85]{,}\hlc[pink!59]{ the}\hlc[pink!4]{ govern}\hlc[pink!99]{or}\hlc[pink!13]{ said}\hlc[pink!95]{.}\hlc[pink!93]{}

+ \hlc[pink!6]{ New}\hlc[pink!48]{ York}\hlc[pink!3]{ state}\hlc[pink!22]{ has}\hlc[pink!33]{ }\hlc[pink!3]{fallen}\hlc[pink!26]{ short}\hlc[pink!98]{ of}\hlc[pink!55]{ the}\hlc[pink!33]{ }\hlc[pink!25]{highest}\hlc[pink!9]{ death}\hlc[pink!89]{ toll}\hlc[pink!54]{ }\hlc[pink!50]{during}\hlc[pink!67]{ the}\hlc[pink!43]{ coronavirus}\hlc[pink!61]{ pandemi}\hlc[pink!99]{c}\hlc[pink!83]{.}\hlc[pink!69]{}

+ \hlc[pink!6]{ New}\hlc[pink!48]{ York}\hlc[pink!3]{ state}\hlc[pink!22]{ has}\hlc[pink!33]{ }\hlc[pink!3]{fallen}\hlc[pink!26]{ below}\hlc[pink!7]{ 100}\hlc[pink!15]{ death}\hlc[pink!12]{s}\hlc[pink!56]{ in}\hlc[pink!50]{ the}\hlc[pink!77]{ last}\hlc[pink!15]{ 24}\hlc[pink!88]{ hours}\hlc[pink!57]{,}\hlc[pink!78]{ }\hlc[pink!80]{a}\hlc[pink!29]{ record}\hlc[pink!50]{ low}\hlc[pink!78]{ for}\hlc[pink!76]{ the}\hlc[pink!45]{ state}\hlc[pink!69]{'}\hlc[pink!99]{s}\hlc[pink!16]{ govern}\hlc[pink!91]{or}\hlc[pink!42]{.}\hlc[pink!85]{}

+ \hlc[pink!6]{ New}\hlc[pink!48]{ York}\hlc[pink!3]{ state}\hlc[pink!22]{ has}\hlc[pink!33]{ }\hlc[pink!3]{fallen}\hlc[pink!26]{ below}\hlc[pink!7]{ 100}\hlc[pink!15]{ death}\hlc[pink!12]{s}\hlc[pink!56]{ from}\hlc[pink!34]{ coronavirus}\hlc[pink!62]{ in}\hlc[pink!53]{ the}\hlc[pink!81]{ last}\hlc[pink!15]{ 24}\hlc[pink!94]{ hours}\hlc[pink!70]{,}\hlc[pink!78]{ }\hlc[pink!82]{according}\hlc[pink!99]{ to}\hlc[pink!59]{ }\hlc[pink!88]{a}\hlc[pink!15]{ govern}\hlc[pink!94]{or}\hlc[pink!68]{.}\hlc[pink!94]{}

+ \hlc[pink!6]{ New}\hlc[pink!48]{ York}\hlc[pink!3]{ state}\hlc[pink!23]{ has}\hlc[pink!33]{ }\hlc[pink!3]{fallen}\hlc[pink!26]{ below}\hlc[pink!7]{ 100}\hlc[pink!15]{ death}\hlc[pink!12]{s}\hlc[pink!56]{ per}\hlc[pink!23]{ day}\hlc[pink!67]{ from}\hlc[pink!28]{ coronavirus}\hlc[pink!56]{ since}\hlc[pink!55]{ the}\hlc[pink!52]{ }\hlc[pink!23]{height}\hlc[pink!73]{ of}\hlc[pink!67]{ the}\hlc[pink!51]{ out}\hlc[pink!99]{break}\hlc[pink!59]{,}\hlc[pink!57]{ the}\hlc[pink!1]{ govern}\hlc[pink!99]{or}\hlc[pink!20]{ said}\hlc[pink!91]{.}\hlc[pink!93]{}
\\
\hline
\end{tabular}
\caption{Prediction examples of the XLSum summarization task.}
\label{tab:example_summarization_2}
\end{table*}

\begin{table*}
\centering
\begin{tabular}{p{0.96\linewidth}}
\hline
\textbf{Query} \\
The ticket from the Euromillions draw on Tuesday, 3 December, worth £40,957,696.60, was bought somewhere in Dorset, the National Lottery said. The operator urged players to "check, double-check and triple-check" their tickets. The winning numbers were 18, 31, 32, 38 and 48 with 4 and 12 as lucky stars. National Lottery spokesman Patrick Lisoire said it was "not very common" to have such a large unclaimed prize. "People are predisposed to checking the tickets they've bought, but for whatever reason - the busy lives we lead, or the lead-up to Christmas - somebody hasn't quite got round to checking their tickets." He said the exact location of where the winning ticket was bought was not being revealed to protect the winner's identity. It is the seventh Euromillions jackpot won in the UK this year. The jackpot must be claimed within six months or the prize money, plus all interest generated, will go to help National Lottery-funded projects across the UK. Tickets for Euromillions are sold in nine countries - the UK, France, Spain, Austria, Belgium, Luxembourg, the Irish Republic, Portugal and Switzerland - with ticket-holders in all those countries trying to win a share of the same jackpot each week. \\
\textbf{Greedy algorithm} \\
+ A jackpot worth £40m has been unclaimed by players in the UK, the National Lottery has said. [x5] \\
\textbf{Baseline top-p (=0.4)} \\
+ \hlc[pink!100]{ A}\hlc[pink!100]{ jackpot}\hlc[pink!100]{ ticket}\hlc[pink!100]{ has}\hlc[pink!100]{ been}\hlc[pink!100]{ un}\hlc[pink!100]{claim}\hlc[pink!100]{ed}\hlc[pink!100]{ from}\hlc[pink!100]{ }\hlc[pink!100]{a}\hlc[pink!100]{ jackpot}\hlc[pink!100]{ draw}\hlc[pink!100]{ in}\hlc[pink!100]{ the}\hlc[pink!100]{ UK}\hlc[pink!100]{,}\hlc[pink!100]{ the}\hlc[pink!100]{ National}\hlc[pink!100]{ L}\hlc[pink!100]{ottery}\hlc[pink!100]{ has}\hlc[pink!100]{ said}\hlc[pink!100]{.}\hlc[pink!100]{}

+ \hlc[pink!100]{ A}\hlc[pink!100]{ jackpot}\hlc[pink!100]{ }\hlc[pink!100]{worth}\hlc[pink!100]{ \textsterling4}\hlc[pink!100]{0,95}\hlc[pink!100]{7}\hlc[pink!100]{,}\hlc[pink!100]{696}\hlc[pink!100]{.60}\hlc[pink!100]{ has}\hlc[pink!100]{ been}\hlc[pink!100]{ un}\hlc[pink!100]{claim}\hlc[pink!100]{ed}\hlc[pink!100]{ by}\hlc[pink!100]{ }\hlc[pink!100]{players}\hlc[pink!100]{ in}\hlc[pink!100]{ the}\hlc[pink!100]{ UK}\hlc[pink!100]{,}\hlc[pink!100]{ the}\hlc[pink!100]{ National}\hlc[pink!100]{ L}\hlc[pink!100]{ottery}\hlc[pink!100]{ has}\hlc[pink!100]{ said}\hlc[pink!100]{.}\hlc[pink!100]{}

+ \hlc[pink!100]{ A}\hlc[pink!100]{ jackpot}\hlc[pink!100]{ }\hlc[pink!100]{worth}\hlc[pink!100]{ \textsterling4}\hlc[pink!100]{m}\hlc[pink!100]{ has}\hlc[pink!100]{ been}\hlc[pink!100]{ un}\hlc[pink!100]{claim}\hlc[pink!100]{ed}\hlc[pink!100]{ by}\hlc[pink!100]{ }\hlc[pink!100]{players}\hlc[pink!100]{ in}\hlc[pink!100]{ the}\hlc[pink!100]{ UK}\hlc[pink!100]{.}\hlc[pink!100]{}

+ \hlc[pink!100]{ A}\hlc[pink!100]{ jackpot}\hlc[pink!100]{ }\hlc[pink!100]{worth}\hlc[pink!100]{ \textsterling}\hlc[pink!100]{4.4}\hlc[pink!100]{m}\hlc[pink!100]{ has}\hlc[pink!100]{ been}\hlc[pink!100]{ un}\hlc[pink!100]{claim}\hlc[pink!100]{ed}\hlc[pink!100]{ by}\hlc[pink!100]{ }\hlc[pink!100]{players}\hlc[pink!100]{ in}\hlc[pink!100]{ the}\hlc[pink!100]{ UK}\hlc[pink!100]{.}\hlc[pink!100]{}

+ \hlc[pink!100]{ A}\hlc[pink!100]{ jackpot}\hlc[pink!100]{ ticket}\hlc[pink!100]{ has}\hlc[pink!100]{ been}\hlc[pink!100]{ un}\hlc[pink!100]{claim}\hlc[pink!100]{ed}\hlc[pink!100]{ in}\hlc[pink!100]{ the}\hlc[pink!100]{ UK}\hlc[pink!100]{ after}\hlc[pink!100]{ }\hlc[pink!100]{a}\hlc[pink!100]{ }\hlc[pink!100]{lotter}\hlc[pink!100]{y}\hlc[pink!100]{ ticket}\hlc[pink!100]{ was}\hlc[pink!100]{ b}\hlc[pink!100]{ought}\hlc[pink!100]{ in}\hlc[pink!100]{ Dor}\hlc[pink!100]{set}\hlc[pink!100]{.}\hlc[pink!100]{}
\\
\textbf{Baseline top-p (=0.7)} \\
+ \hlc[pink!100]{ A}\hlc[pink!100]{ jackpot}\hlc[pink!100]{ in}\hlc[pink!100]{ the}\hlc[pink!100]{ UK}\hlc[pink!100]{ has}\hlc[pink!100]{ been}\hlc[pink!100]{ un}\hlc[pink!100]{claim}\hlc[pink!100]{ed}\hlc[pink!100]{,}\hlc[pink!100]{ }\hlc[pink!100]{according}\hlc[pink!100]{ to}\hlc[pink!100]{ the}\hlc[pink!100]{ National}\hlc[pink!100]{ L}\hlc[pink!100]{ottery}\hlc[pink!100]{.}\hlc[pink!100]{}

+ \hlc[pink!100]{ A}\hlc[pink!100]{ }\hlc[pink!100]{winning}\hlc[pink!100]{ ticket}\hlc[pink!100]{ from}\hlc[pink!100]{ the}\hlc[pink!100]{ UK}\hlc[pink!100]{ Euro}\hlc[pink!100]{millions}\hlc[pink!100]{ jackpot}\hlc[pink!100]{ has}\hlc[pink!100]{ been}\hlc[pink!100]{ un}\hlc[pink!100]{claim}\hlc[pink!100]{ed}\hlc[pink!100]{,}\hlc[pink!100]{ }\hlc[pink!100]{according}\hlc[pink!100]{ to}\hlc[pink!100]{ the}\hlc[pink!100]{ National}\hlc[pink!100]{ L}\hlc[pink!100]{ottery}\hlc[pink!100]{.}\hlc[pink!100]{}

+ \hlc[pink!100]{ A}\hlc[pink!100]{ jackpot}\hlc[pink!100]{ from}\hlc[pink!100]{ the}\hlc[pink!100]{ UK}\hlc[pink!100]{'}\hlc[pink!100]{s}\hlc[pink!100]{ }\hlc[pink!100]{largest}\hlc[pink!100]{ }\hlc[pink!100]{lotter}\hlc[pink!100]{y}\hlc[pink!100]{ has}\hlc[pink!100]{ been}\hlc[pink!100]{ un}\hlc[pink!100]{claim}\hlc[pink!100]{ed}\hlc[pink!100]{,}\hlc[pink!100]{ the}\hlc[pink!100]{ operator}\hlc[pink!100]{ has}\hlc[pink!100]{ said}\hlc[pink!100]{.}\hlc[pink!100]{}

+ \hlc[pink!100]{ The}\hlc[pink!100]{ jackpot}\hlc[pink!100]{ from}\hlc[pink!100]{ the}\hlc[pink!100]{ Euro}\hlc[pink!100]{millions}\hlc[pink!100]{ jackpot}\hlc[pink!100]{ draw}\hlc[pink!100]{ has}\hlc[pink!100]{ been}\hlc[pink!100]{ un}\hlc[pink!100]{claim}\hlc[pink!100]{ed}\hlc[pink!100]{ after}\hlc[pink!100]{ }\hlc[pink!100]{a}\hlc[pink!100]{ }\hlc[pink!100]{winning}\hlc[pink!100]{ ticket}\hlc[pink!100]{ was}\hlc[pink!100]{ sold}\hlc[pink!100]{ in}\hlc[pink!100]{ the}\hlc[pink!100]{ UK}\hlc[pink!100]{,}\hlc[pink!100]{ the}\hlc[pink!100]{ National}\hlc[pink!100]{ L}\hlc[pink!100]{ottery}\hlc[pink!100]{ has}\hlc[pink!100]{ said}\hlc[pink!100]{.}\hlc[pink!100]{}

+ \hlc[pink!100]{ The}\hlc[pink!100]{ jackpot}\hlc[pink!100]{ }\hlc[pink!100]{earned}\hlc[pink!100]{ by}\hlc[pink!100]{ }\hlc[pink!100]{a}\hlc[pink!100]{ ticket}\hlc[pink!100]{ }\hlc[pink!100]{holder}\hlc[pink!100]{ has}\hlc[pink!100]{ been}\hlc[pink!100]{ }\hlc[pink!100]{claim}\hlc[pink!100]{ed}\hlc[pink!100]{ }\hlc[pink!100]{during}\hlc[pink!100]{ the}\hlc[pink!100]{ Christmas}\hlc[pink!100]{ period}\hlc[pink!100]{.}\hlc[pink!100]{}
\\
\textbf{Guided top-p (=1.0)} \\
+ \hlc[pink!12]{ A}\hlc[pink!0]{ jackpot}\hlc[pink!28]{ }\hlc[pink!34]{worth}\hlc[pink!28]{ \textsterling4}\hlc[pink!6]{0,95}\hlc[pink!19]{7}\hlc[pink!23]{,}\hlc[pink!1]{696}\hlc[pink!0]{.60}\hlc[pink!13]{ has}\hlc[pink!88]{ been}\hlc[pink!13]{ un}\hlc[pink!0]{claim}\hlc[pink!99]{ed}\hlc[pink!76]{ in}\hlc[pink!54]{ the}\hlc[pink!42]{ UK}\hlc[pink!52]{,}\hlc[pink!41]{ the}\hlc[pink!1]{ National}\hlc[pink!3]{ L}\hlc[pink!99]{ottery}\hlc[pink!2]{ has}\hlc[pink!34]{ said}\hlc[pink!98]{.}\hlc[pink!79]{}

+ \hlc[pink!12]{ A}\hlc[pink!0]{ jackpot}\hlc[pink!28]{ }\hlc[pink!34]{worth}\hlc[pink!28]{ \textsterling}\hlc[pink!60]{40}\hlc[pink!98]{m}\hlc[pink!73]{ has}\hlc[pink!87]{ been}\hlc[pink!18]{ un}\hlc[pink!0]{claim}\hlc[pink!99]{ed}\hlc[pink!76]{ in}\hlc[pink!49]{ the}\hlc[pink!42]{ UK}\hlc[pink!69]{,}\hlc[pink!43]{ the}\hlc[pink!2]{ National}\hlc[pink!4]{ L}\hlc[pink!99]{ottery}\hlc[pink!4]{ has}\hlc[pink!22]{ said}\hlc[pink!98]{.}\hlc[pink!81]{}

+ \hlc[pink!12]{ A}\hlc[pink!0]{ jackpot}\hlc[pink!28]{ }\hlc[pink!34]{worth}\hlc[pink!28]{ \textsterling4}\hlc[pink!7]{0,95}\hlc[pink!19]{7}\hlc[pink!23]{,}\hlc[pink!1]{696}\hlc[pink!0]{.60}\hlc[pink!13]{ has}\hlc[pink!88]{ been}\hlc[pink!13]{ un}\hlc[pink!0]{claim}\hlc[pink!99]{ed}\hlc[pink!76]{ after}\hlc[pink!70]{ }\hlc[pink!82]{a}\hlc[pink!5]{ ticket}\hlc[pink!26]{ was}\hlc[pink!15]{ b}\hlc[pink!97]{ought}\hlc[pink!42]{ in}\hlc[pink!20]{ Dor}\hlc[pink!97]{set}\hlc[pink!69]{,}\hlc[pink!39]{ the}\hlc[pink!0]{ National}\hlc[pink!3]{ L}\hlc[pink!99]{ottery}\hlc[pink!0]{ said}\hlc[pink!90]{.}\hlc[pink!84]{}

+ \hlc[pink!12]{ A}\hlc[pink!0]{ jackpot}\hlc[pink!28]{ }\hlc[pink!34]{worth}\hlc[pink!28]{ \textsterling}\hlc[pink!60]{4.8}\hlc[pink!86]{m}\hlc[pink!75]{ has}\hlc[pink!91]{ been}\hlc[pink!17]{ un}\hlc[pink!0]{claim}\hlc[pink!99]{ed}\hlc[pink!77]{ after}\hlc[pink!71]{ }\hlc[pink!83]{a}\hlc[pink!5]{ ticket}\hlc[pink!29]{ was}\hlc[pink!17]{ b}\hlc[pink!98]{ought}\hlc[pink!45]{ in}\hlc[pink!20]{ the}\hlc[pink!21]{ UK}\hlc[pink!46]{,}\hlc[pink!32]{ the}\hlc[pink!1]{ National}\hlc[pink!3]{ L}\hlc[pink!99]{ottery}\hlc[pink!1]{ said}\hlc[pink!97]{.}\hlc[pink!83]{}

+ \hlc[pink!12]{ A}\hlc[pink!0]{ jackpot}\hlc[pink!28]{ }\hlc[pink!34]{worth}\hlc[pink!28]{ \textsterling}\hlc[pink!60]{400,000}\hlc[pink!78]{ has}\hlc[pink!89]{ been}\hlc[pink!15]{ un}\hlc[pink!0]{claim}\hlc[pink!99]{ed}\hlc[pink!75]{ from}\hlc[pink!58]{ }\hlc[pink!26]{a}\hlc[pink!5]{ ticket}\hlc[pink!18]{ b}\hlc[pink!83]{ought}\hlc[pink!64]{ in}\hlc[pink!17]{ Dor}\hlc[pink!96]{set}\hlc[pink!58]{.}\hlc[pink!72]{}
\\
\hline
\end{tabular}
\caption{Prediction examples of the XLSum summarization task.}
\label{tab:example_summarization_3}
\end{table*}